\documentclass{jfpc}
 
\usepackage[latin9]{inputenc} 
\usepackage{amssymb}
\usepackage{graphicx}
\usepackage{url}
\urldef{\mailsa}\path|Mohammed.Bekkouche@unice.fr|

\usepackage[babel=true]{csquotes}

\usepackage{lscape} 
\usepackage{wrapfig}

\usepackage[usenames,dvipsnames]{color}

\usepackage{algorithmic,algorithm2e,float}

\usepackage{listings}
\usepackage{amssymb}
\usepackage{amsmath}
\usepackage{amssymb}
\usepackage{amsfonts}
\usepackage{csquotes}
\usepackage{ulem}
\usepackage{hhline}

\usepackage[table]{xcolor}
\usepackage{longtable}
\usepackage{multirow}
\usepackage{pgfplots}

\usepackage{color}
\usepackage{tikz}
\usetikzlibrary{positioning}
\usetikzlibrary{shapes.geometric}
\usetikzlibrary{snakes}
\usetikzlibrary{arrows}
\definecolor{dkgreen}{rgb}{0,0.6,0}
\definecolor{gray}{rgb}{0.5,0.5,0.5}
\definecolor{mauve}{rgb}{0.58,0,0.82}

\annee{2015}
\titre{Exploration de la scalabilité de {\tt LocFaults}}

\auteurs{Mohammed Bekkouche} 

\institutions{Univ. Nice Sophia Antipolis, CNRS, I3S, UMR 7271, 06900 Sophia Antipolis, France
}

\mels{Mohammed.Bekkouche@unice.fr}

\begin{document}

\creationEntete

\begin{resume} % environnement pour le résumé en français

Un vérificateur de modèle peut produire une trace de contre-exemple, pour un programme erroné, qui est souvent longue et difficile à comprendre. En général, la partie qui concerne les boucles est la plus importante parmi les instructions de cette trace. Ce qui rend la localisation d'erreurs dans les boucles cruciale, pour analyser les erreurs dans le programme en global. Dans ce papier, nous explorons les capacités de la scalabilité de {\tt LocFaults}, notre approche  de localisation d'erreurs exploitant les chemins du CFG(Controle Flow Graph) à partir d'un contre-exemple pour calculer les DCMs(Déviations de Correction Minimales), ainsi les MCSs(Minimal Correction Subsets) à partir de chaque DCM. Nous présentons les temps de notre approche sur des programmes avec boucles \textit{While} dépliées $b$ fois, et un nombre de conditions déviées allant de $0$ à $n$. Nos résultats préliminaires montrent que les temps de notre approche, basée sur les contraintes et dirigée par les flots, sont meilleurs par rapport à {\tt BugAssist} qui se base sur SAT et transforme la totalité du programme en une formule booléenne, et de plus l'information fournie par {\tt LocFaults} est plus expressive pour l'utilisateur.

\end{resume}

\begin{abstract} % environnement pour le résumé en anglais
A model checker can produce a trace of counterexample, for a erroneous program, which is often long and difficult to understand. In general, the part about the loops is the largest among the instructions in this trace. This makes the location of errors in loops critical, to analyze errors in the overall program. In this paper, we explore the scalability capabilities of {\tt LocFaults}, our error localization approach  exploiting paths of CFG(Control Flow Graph) from a counterexample to calculate the MCDs (Minimal Correction Deviations), and MCSs (Minimal Correction Subsets) from each MCD found. We present the times of our approach on programs with \textit{While}-loops unfolded $b$ times, and a number of diverted conditions ranging from $0$ to $n$. Our preliminary results show that the times of our approach, constraint-based and flow-driven, are better compared to {\tt BugAssist} which is based on SAT and transforms the entire program to a Boolean formula, although the information provided by {\tt LocFaults} is more expressive for the user.

\end{abstract}

\section{Introduction}
Les erreurs dans un programme sont inévitables, elles peuvent nuire à son bon fonctionnement et avoir des conséquences financières extrêmement graves et présenter une menace pour le bien-être humain~\cite{listofsoftwarebugs}. Le lien suivant~\cite{bugstories} cite des histoires récentes de bugs logiciels. Conséquemment, le processus de débogage (la détection, la localisation et la correction d'erreurs) est essentiel. La localisation d'erreurs est l'étape qui coûte le plus. Elle consiste à identifier l'emplacement exact des instructions suspectes~\cite{wong2009survey} afin d'aider l'utilisateur à comprendre pourquoi le programme a échoué, ce qui lui facilite la tâche de la correction des erreurs. En effet, quand un programme P est non conforme vis-à-vis de sa spécification (P contient des erreurs), un vérificateur de modèle peut produire une trace d'un contre-exemple, qui est souvent longue et difficile à comprendre même pour les programmeurs expérimentés. Pour résoudre ce problème, nous avons proposé une approche~\cite{locfaults2} (nommée {\tt LocFaults}) à base de contraintes qui explore les chemins du CFG(Control Flow Graph) du programme à partir du contre-exemple, pour calculer les sous-ensembles minimaux permettant de restaurer la conformité du programme vis-à-vis de sa postcondition. Assurer que notre méthode soit hautement scalable pour faire face à l'énorme complexité des systèmes logiciels est un critère important pour sa qualité~\cite{d2008survey}.

Dans ce papier, nous explorons le passage à l'échelle de {\tt LocFaults} sur des programmes avec boucles \textit{While} dépliées $b$ fois, et un nombre de conditions déviées allant de $0$ à $3$.  

L'idée de notre approche est de réduire le problème de la localisation d'erreurs vers celui qui consiste à calculer un ensemble minimal qui explique pourquoi un CSP (Constraint Satisfaction Problem) est infaisable. Le CSP représente l'union des contraintes du contre-exemple, du programme et de l'assertion violée. L'ensemble calculé peut être un MCS (Minimal Correction Subset) ou MUS (Minimal Unsatisfiable Subset). En général, tester la faisabilité d'un CSP sur un domaine fini est un problème NP-Complet (intraitable)   \footnote{Si ce problème pouvait être résolu en temps polynomial, alors tous les problèmes NP-Complet le seraient aussi.}, la classe des problèmes les plus difficiles de la classe NP. Cela veut dire, expliquer l'infaisabilité dans un CSP est aussi dur, voire plus (on peut classer le problème comme NP-Difficile). {\tt BugAssist}~\cite{mrbugassist2011}~\cite{mrbugassist2011tool} est une méthode de localisation d'erreurs qui utilise un solveur Max-SAT pour calculer la fusion des MCSs de la formule Booléenne du programme en entier avec le contre-exemple. Elle devient inefficace pour les programmes de grande taille. {\tt LocFaults} travaille aussi à partir d'un contre-exemple pour calculer les MCSs. La contribution de notre approche par rapport à {\tt BugAssist} peut se résumer dans les points suivants : 
\begin{itemize}
\item[*] Nous ne transformons pas la totalité du programme en un système de contraintes, mais nous utilisons le CFG du programme pour collecter les contraintes du chemin du contre-exemple et des chemins dérivés de ce dernier, en supposant qu'au plus $k$ instructions conditionnelles sont susceptibles de contenir les erreurs. Nous calculons les MCSs uniquement sur le chemin du contre-exemple et les chemins qui corrigent le programme ;
\item[*] Nous ne traduisons pas les instructions du programme en une formule SAT, mais plutôt en contraintes numériques qui vont être manipulées par des solveurs de contraintes ;
\item[*] Nous n'utilisons pas des solveurs MaxSAT comme boîtes noires, mais plutôt un algorithme générique pour calculer les MCSs par l'usage d'un solveur de contraintes ;
\item[*] Nous bornons la taille des MCSs générés et le nombre de conditions déviées ;
\item[*] Nous pouvons faire collaborer plusieurs solveurs durant le processus de localisation et prendre celui le plus performant selon la catégorie du CSP construit. Exemple, si le CSP du chemin détecté est du type linéaire sur les entiers, nous faisons appel à un solveur MIP (Mixed Integer Programming) ; s'il est non linéaire, nous utilisons un solveur CP (Constraint Programming) ou aussi MINLP (Mixed Integer Nonlinear Programming).
\end{itemize}
   
Notre expérience pratique a montré que toutes ces restrictions et distinctions ont permis à {\tt LocFaults} d'être plus rapide et plus expressif. 

Le papier est organisé comme suit. La section 2 introduit la définition d'un MUS et MCS. Dans la section 3, nous définirons le problème $\leq k$-DCM. Nous expliquons une contribution du papier pour le traitement des boucles erronées, notamment le bug \textit{Off-by-one}, dans la section 4. Une brève description de notre algorithme {\tt LocFaults} est fournie dans la section 5. L'évaluation expérimentale est présentée dans la section 6. La section 7 parle de la conclusion et de nos travaux futurs.

\section{Définitions}
Dans cette section, nous introduirons la définition d'un IIS/MUS et MCS. 
\paragraph{CSP}
Un $CSP$(Constraint Satisfaction Problem) $P$ est un triplet $<X,D,C>$ tel que :
\begin{itemize}
\item[*] $X$ un ensemble de n variables $x_{1}, x_{2}, ..., x_{n}$.
\item[*] $D$ le n-uplet $<D_{x_{1}}, D_{x_{2}}, ..., D_{x_{n}}>$. L'ensemble $D_{x_{i}}$ contient les valeurs de la variable $x_{i}$.   
\item[*] $C$=$\{c_{1}, c_{2}, ..., c_{n}\}$ est l'ensemble des contraintes.
\end{itemize}

%Soit $\Delta=$ $D_{x_{1}} \times D_{x_{2}} \times ... \times D_{x_{n}}$ l'ensemble des n-uplets qui représente toutes les combinaisons possibles que peuvent prendre les $n$ variables de $X$ dans $D$. 
Une \textit{solution} pour $P$ est une instanciation des variables $\cal{I}$ $\in$ $D$ qui satisfait toutes les contraintes dans $C$. $P$ est infaisable s'il ne dispose pas de solutions. Un sous-ensemble de contraintes $C'$ dans $C$ est dit aussi infaisable pour la même raison sauf qu'ici on se limite à l'ensemble des contraintes dans $C'$.\newline 
On note par :
\begin{itemize}
\item $Sol(<X,C',D>)=\emptyset$, pour spécifier que $C'$ n'a pas de solutions, et donc il est infaisable.  
\item $Sol(<X,C',D>) \neq \emptyset$, pour spécifier que $C'$ dispose d'au moins une solution, et donc il est faisable. 
\end{itemize}
%\begin{definition}[Définition 2 (La fonction Sol).]
%En partant de notre $CSP$ défini $\cal{P}$, on définit la fonction $Sol$ de $C \times D$ dans $\Delta$. La fonction $Sol$ prend en entrée un système de contraintes $C$ et un domaine $D$, formulée $Sol(C, D)$, et fournit l'ensemble des solutions des contraintes $C$ où les variables prennent leurs valeurs dans $D$.  
%\end{definition}

On dit que $P$ est en forme \textit{linéaire} et on note LP(Linear Program) ssi toutes les contraintes dans $C$ sont des équations/inégalités linéaires, il est \textit{continu} si le domaine de toutes les variables est celui des réels. Si au moins une des variables dans $X$ est du type entier ou binaire (cas spécial d'un entier), et les contraintes sont linéaires, $P$ est dit un programme \textit{linéaire mixte} MIP(Mixed-integer linear program).
Si les contraintes sont non-linéaires, on dit que $P$ est un programme \textit{non linéaire} NLP(NonLinear Program).    

Soit $P=<X,D,C>$ un $CSP$ infaisable, on définit pour $P$ :
\paragraph{IS} 
Un IS(Inconsistent Set) est un sous-ensemble de contraintes infaisable dans l'ensemble de contraintes infaisable $C$. 
%\enquote{Where the subset is infeasible, but is reducible, it is simply called an infeasible subset (IS) of constraints} (Chinneck~\cite{chinneck2007feasibility}, Page.93).\\
$C'$ est un IS ssi :
\begin{itemize}
\item[*] $C'$ $\subseteq$ $C$. 
\item[*] $Sol(<X,C',D>)=\emptyset$.
\end{itemize}

\paragraph{IIS ou MUS}
Un IIS(Irreducible Inconsistent Set) ou MUS (Minimal Unsatisfiable Subset) est un sous-ensemble de contraintes infaisable de $C$, et tous ses sous-ensembles stricts sont faisables. 
%\enquote{An IIS has this property: it is itself infeasible, but any proper subset is feasible} (Chinneck~\cite{chinneck2007feasibility}, P.93). \enquote{an irreducible infeasible subset (IIS) of the constraints, i.e. a (small) subset of constraints that is itself infeasible, but becomes feasible if one or more constraints is removed}(Chinneck~\cite{chinneck2007feasibility}, Page.4).\\
$C'$ est un IIS ssi :
\begin{itemize}
\item[*] $C'$ est un IS.
\item[*] $\forall$ $C''$ $\subset$ $C'$.$Sol(<X,C'',D>)\neq\emptyset$, (chacune de ses parties contribue à l'infaisabilité), $C'$ est dit irréductible.
\end{itemize}

\paragraph{MCS}
$C'$ est un MCS(Minimal Correction Set) ssi :
\begin{itemize}
\item[*] $C'$ $\subseteq$ $C$.
\item[*] $Sol(<X,C \backslash C',D>)\neq\emptyset$.
\item[*] $\nexists$ $C''$ $\subset$ $C'$ tel que $Sol(<X,C \backslash C'',D>)\neq\emptyset$.
\end{itemize}

%\begin{definition}[Définition MSS]
%\textbf{\newline}
%$C'$ est un MSS(Maximal Satisfiable Set) ssi :
%\begin{itemize}
%\item[*] $C'$ $\subseteq$ $C$.
%\item[*] $Sol(<X,C',D>)\neq\emptyset$.
%\item[*] $\nexists$ $C''$ $\subset$ $C \backslash C'$ tel que $Sol(<X,C' \cup C'',D>)\neq\emptyset$.
%\end{itemize}
%\end{definition}

\section{Le problème $\leq k$-DCM}
Étant donné un programme erroné modélisé en un CFG\footnote{Nous utilisons la transformation en forme DSA~\cite{dsa2005} qui assure que chaque variable est affectée une seule fois sur chaque chemin du CFG.} $G=(C,A,E)$: $C$ est l'ensemble des n\oe{}uds conditionnels ; $A$ est l'ensemble des blocs d'affectation ; $E$ est l'ensemble des arcs, et un contre-exemple. Une DCM (\textit{Déviation de Correction Minimale}) est un ensemble $D$ $\subseteq$ $C$ telle que la propagation du contre-exemple sur l'ensemble des instructions de G à partir de la racine, tout en ayant nié chaque condition\footnote{On nie la condition afin de prendre la branche opposée à celle où on devait aller.} dans $D$, permet en sortie de satisfaire la postcondition. Elle est dite minimale (ou irréductible) dans le sens où aucun élément ne peut être retiré de $D$ sans que celle-ci ne perde cette propriété. En d'autres termes, $D$ est une correction minimale du programme dans l'ensemble des conditions. La taille d'une déviation minimale est son cardinal. Le problème \textit{$\leq k$-DCM} consiste à trouver toutes les DCMs de taille inférieure ou égale à $k$.

Exemple, le CFG du programme AbsMinus (voir fig. 2) possède une déviation minimale de taille $1$ pour le contre-exemple $\{i=0,j=1\}$. Certes, la déviation \{$i_0 \leq j_0$,$k_1=1 \land i_0 \neq j_0$\} permet de corriger le programme, mais elle n'est pas minimale ; la seule déviation minimale pour ce programme est \{$k_1=1 \land i_0 \neq j_0$\}.
 
\begin{figure}[!h]
\begin{minipage}{1cm}
\end{minipage} \hfill
\begin{minipage}[b]{.45\linewidth}
\begin{lstlisting}{}
class AbsMinus {	
/*@ ensures 
  @ ((i<j)==>(\result==j-i))&& 
  @ ((i>=j)==>(\result==i-j));*/
  int AbsMinus (int i, int j){
   int result;
   int k = 0;
	 if (i <= j) {
	  k = k+2;//error: should be k=k+1
	 }
   if (k == 1 && i != j) {
	  result = j-i; 		
	 }  
	 else {
	  result = i-j;
	 }
  }
}
\end{lstlisting}
\vspace{-1cm}
\caption{{\scriptsize Le programme AbsMinus}}
\label{AbsMinus}
\end{minipage}
\begin{minipage}[b]{.45\linewidth}
\begin{tiny}  
\begin{tikzpicture}[auto,node distance=2cm,/.style={font=\sffamily\footnotesize}]
\tikzstyle{decision} = [scale=0.6,diamond, draw=black, thick,text width=3em, text badly centered, inner sep=1pt]
\tikzstyle{block} = [scale=0.6,rectangle, draw=black, thick, 
text width=5.5em, text centered, rounded corners, minimum height=0.8em,text height=0.5em]
\tikzstyle{bloc} = [scale=0.6,rectangle, draw=black, thick, text centered, minimum height=0.8em,text height=0.5em]
\tikzstyle{line} = [scale=0.6,draw, thick, -latex',shorten >=1pt];
\tikzstyle{cloud} = [scale=0.6,thick, ellipse,draw=black, minimum height=2em];
\matrix [column sep=0.5mm,row sep=2mm,ampersand replacement=\&]
{
% row 1
%\& \node [cloud] (1) {$(i_0==0)$ $\land$ $(j_0==1)$}; \& \\
% row 2
%& \node [block] (2) {$r_0=0$}; & \\
% row 3
\& \node [block] (4) {$k_0=0$}; \& \\
% row 4
\& \node [decision] (5) {$i_0$ $\leq$ $j_0$}; \& \\
% row 5
\node [block] (3) {$k_1=k_0+2$};
\& \node[draw=red] (11) {Error}; \& \node [block] (6) {$k_1=k_0$}; \\
% row 6
\& \node [decision,text width=6.3em] (7) {$k_1=1 \land i_0!=j_0$}; \&  \\ 
% row 7
\node [block] (8) {$r_1=j_0-i_0$};
\&  \& \node [block] (9) {$r_1=i_0-j_0$}; \\
% row 8
\& \node [bloc] (10) {POST:$\{r_1==|i-j|\}$}; \& \\
};
\tikzstyle{every path}=[line,distance=1cm]
%\path (1) -- (4);
\path (4) -- (5);
\path (5.west) -- node [left] {If} (3.north);
\path (5.east) -- node [right] {Else} (6.north);
\path (3.south) -- (7);
\path (6.south) -- (7);
\path (7.west) -- node [left] {If} (8.north);
\path (7.east) -- node [right] {Else} (9.north);
\path (8.south) -- (10.west);
\path (9.south) -- (10.east);
\path[snake=zigzag] (11.west) -- (3.east);
\end{tikzpicture} 
\vspace{-0.5cm}
\label{cfgAbsMinus}
\caption{{\scriptsize Le CFG DSA de AbsMinus}}
\end{tiny}
\end{minipage}
\end{figure}

\begin{figure}[!h]
\begin{minipage}{2cm}
\end{minipage} \hfill
\begin{minipage}{4cm}
\begin{tiny}
\begin{tikzpicture}[auto,node distance=2cm,/.style={font=\sffamily\footnotesize}]
\tikzstyle{decision} = [scale=0.6,diamond, draw=black, thick,text width=3em, text badly centered, inner sep=1pt]
\tikzstyle{block} = [scale=0.6,rectangle, draw=black, thick, 
text width=5.5em, text centered, rounded corners, minimum height=0.8em,text height=0.5em]
\tikzstyle{bloc} = [scale=0.6,rectangle, draw=black, thick, text centered, minimum height=0.8em,text height=0.5em]
\tikzstyle{line} = [scale=0.6,draw, thick, -latex',shorten >=1pt];
\tikzstyle{cloud} = [scale=0.6,thick, ellipse,draw=black, minimum height=2em];
\matrix [column sep=0.5mm,row sep=2mm,ampersand replacement=\&]
{
% row 1
\& \node [bloc,draw=red] (1) {$\{(i_0==0)$ $\land$ $(j_0==1)\}$}; \& \\
% row 2
%& \node [block] (2) {$r_0=0$}; & \\
% row 3
\& \node [block,draw=red] (4) {$k_0=0$}; \& \\
% row 4
\& \node [decision,draw=red] (5) {$i_0$ $\leq$ $j_0$}; \& \\
% row 5
\node [block,draw=red] (3) {$k_1=k_0+2$};
\&  \& \node [block] (6) {$k_1=k_0$}; \\
% row 6
\& \node [decision,text width=6.3em,draw=red] (7) {$k_1=1 \land i_0!=j_0$}; \&s  \\ 
% row 7
\node [block] (8) {$r_1=j_0-i_0$};
\&  \& \node [block,draw=red] (9) {$r_1=i_0-j_0$}; \\
% row 8
\& \node [bloc,draw=red] (10) {$\{r_1==|i-j|\}$}; \& \\
};
\tikzstyle{every path}=[line,distance=1cm]
\path[draw=red] (1) -- (4);
\path[draw=red] (4) -- (5);
\path[draw=red] (5.west) -- node [left] {If} (3.north);
\path (5.east) -- node [right] {Else} (6.north);
\path[draw=red] (3.south) -- (7);
\path (6.south) -- (7);
\path (7.west) -- node [left] {If} (8.north);
\path[draw=red] (7.east) -- node [right] {Else} (9.north);
\path (8.south) -- (10.west);
\path[draw=red] (9.south) -- (10.east);
\end{tikzpicture}
 \vspace{-0.5cm}
\caption{{\scriptsize Le chemin du contre-exemple}} 
\label{cfgAbsMinus1} 
\end{tiny}
\end{minipage} \hfill
\begin{minipage}[c]{4cm}
\begin{tiny}
\begin{tikzpicture}[auto,node distance=2cm,/.style={font=\sffamily\footnotesize}]
\tikzstyle{decision} = [scale=0.6,diamond, draw=black, thick,text width=3em, text badly centered, inner sep=1pt]
\tikzstyle{block} = [scale=0.6,rectangle, draw=black, thick, 
text width=5.5em, text centered, rounded corners, minimum height=0.8em,text height=0.5em]
\tikzstyle{bloc} = [scale=0.6,rectangle, draw=black, thick, text centered, minimum height=0.8em,text height=0.5em]
\tikzstyle{line} = [scale=0.6,draw, thick, -latex',shorten >=1pt];
\tikzstyle{cloud} = [scale=0.6,thick, ellipse,draw=black, minimum height=2em];
\matrix [column sep=0.5mm,row sep=2mm,ampersand replacement=\&]
{
% row 1
\& \node [bloc,draw=red] (1) {$\{(i_0==0)$ $\land$ $(j_0==1)\}$}; \& \\
% row 2
%& \node [block] (2) {$r_0=0$}; & \\
% row 3
\& \node [block,draw=red] (4) {$k_0=0$}; \& \\
% row 4
\& \node [decision,draw=red,fill=red!5] (5) {$i_0$ $\leq$ $j_0$}; \& \\
% row 5
\node [block] (3) {$k_1=k_0+2$};
\&  \& \node [block,draw=red] (6) {$k_1=k_0$}; \\
% row 6
\& \node [decision,text width=6.3em,draw=red] (7) {$k_1=1 \land i_0!=j_0$}; \&s  \\ 
% row 7
\node [block] (8) {$r_1=j_0-i_0$};
\&  \& \node [block,draw=red] (9) {$r_1=i_0-j_0$}; \\
% row 8
\& \node [bloc,draw=red] (10) {$\{r_1==|i-j|\}$ is UNSAT}; \& \\
};
\tikzstyle{every path}=[line,distance=1cm]
\path[draw=red] (1) -- (4);
\path[draw=red] (4) -- (5);
\path (5.west) -- node [left] {If} (3.north);
\path[draw=red] (5.east) -- node [right] {Else} (6.north);
\path (3.south) -- (7);
\path[draw=red] (6.south) -- (7);
\path (7.west) -- node [left] {If} (8.north);
\path[draw=red] (7.east) -- node [right] {Else} (9.north);
\path (8.south) -- (10.west);
\path[draw=red] (9.south) -- (10.east);
\end{tikzpicture}
\vspace{-0.5cm}
\caption{{\scriptsize Le chemin obtenu en déviant la condition $i_0$ $\leq$ $j_0$}} 
\label{cfgAbsMinus2}  
\end{tiny}
\end{minipage} \hfill
\begin{minipage}{2cm}
\end{minipage} \hfill
\end{figure}

\begin{figure}[!h]
\begin{minipage}{2cm}
\end{minipage} \hfill
\begin{minipage}[c]{4cm}
\begin{tiny}
\begin{tikzpicture}[auto,node distance=2cm,/.style={font=\sffamily\footnotesize}]
\tikzstyle{decision} = [scale=0.6,diamond, draw=black, thick,text width=3em, text badly centered, inner sep=1pt]
\tikzstyle{block} = [scale=0.6,rectangle, draw=black, thick, 
text width=5.5em, text centered, rounded corners, minimum height=0.8em,text height=0.5em]
\tikzstyle{bloc} = [scale=0.6,rectangle, draw=black, thick, text centered, minimum height=0.8em,text height=0.5em]
\tikzstyle{line} = [scale=0.6,draw, thick, -latex',shorten >=1pt];
\tikzstyle{cloud} = [scale=0.6,thick, ellipse,draw=black, minimum height=2em];
\matrix [column sep=0.5mm,row sep=2mm,ampersand replacement=\&]
{
% row 1
\& \node [bloc,draw=red] (1) {$\{(i_0==0)$ $\land$ $(j_0==1)\}$}; \& \\
% row 2
%& \node [block] (2) {$r_0=0$}; & \\
% row 3
\& \node [block,draw=red] (4) {$k_0=0$}; \& \\
% row 4
\& \node [decision,draw=red] (5) {$i_0$ $\leq$ $j_0$}; \& \\
% row 5
\node [block,draw=red] (3) {$k_1=k_0+2$};
\&  \& \node [block] (6) {$k_1=k_0$}; \\
% row 6
\& \node [decision,text width=6.3em,draw=red,fill=red!5] (7) {$k_1=1 \land i_0!=j_0$}; \&s  \\ 
% row 7
\node [block,draw=red] (8) {$r_1=j_0-i_0$};
\&  \& \node [block] (9) {$r_1=i_0-j_0$}; \\
% row 8
\& \node [bloc,draw=red] (10) {\textbf{$\{r_1==|i-j|\}$ {is SAT}}}; \& \\
};
\tikzstyle{every path}=[line,distance=1cm]
\path[draw=red] (1) -- (4);
\path[draw=red] (4) -- (5);
\path[draw=red] (5.west) -- node [left] {If} (3.north);
\path (5.east) -- node [right] {Else} (6.north);
\path[draw=red] (3.south) -- (7);
\path (6.south) -- (7);
\path[draw=red] (7.west) -- node [left] {If} (8.north);
\path (7.east) -- node [right] {Else} (9.north);
\path[draw=red] (8.south) -- (10.west);
\path (9.south) -- (10.east);
\end{tikzpicture}
\vspace{-0.5cm}
\caption{{\scriptsize Le chemin en déviant la condition $k_1=1 \land i_0!=j_0$}} 
\label{cfgAbsMinus3}
\end{tiny} 
\end{minipage} \hfill
\begin{minipage}{4cm}
\begin{tiny}
\begin{tikzpicture}[auto,node distance=2cm,/.style={font=\sffamily\footnotesize}]
\tikzstyle{decision} = [scale=0.6,diamond, draw=black, thick,text width=3em, text badly centered, inner sep=1pt]
\tikzstyle{block} = [scale=0.6,rectangle, draw=black, thick, 
text width=5.5em, text centered, rounded corners, minimum height=0.8em,text height=0.5em]
\tikzstyle{bloc} = [scale=0.6,rectangle, draw=black, thick, text centered, minimum height=0.8em,text height=0.5em]
\tikzstyle{line} = [scale=0.6,draw, thick, -latex',shorten >=1pt];
\tikzstyle{cloud} = [scale=0.6,thick, ellipse,draw=black, minimum height=2em];
\matrix [column sep=0.5mm,row sep=2mm,ampersand replacement=\&]
{
% row 1
\& \node [bloc,draw=red] (1) {$\{(i_0==0)$ $\land$ $(j_0==1)\}$}; \& \\
% row 2
%& \node [block] (2) {$r_0=0$}; & \\
% row 3
\& \node [block,draw=red] (4) {$k_0=0$}; \& \\
% row 4
\& \node [decision,draw=red,fill=red!5] (5) {$i_0$ $\leq$ $j_0$}; \& \\
% row 5
\node [block] (3) {$k_1=k_0+2$};
\&  \& \node [block,draw=red] (6) {$k_1=k_0$}; \\
% row 6
\& \node [decision,text width=6.3em,draw=red,fill=red!5] (7) {$k_1=1 \land i_0!=j_0$}; \&s  \\ 
% row 7
\node [block,draw=red] (8) {$r_1=j_0-i_0$};
\&  \& \node [block] (9) {$r_1=i_0-j_0$}; \\
% row 8
\& \node [bloc,draw=red] (10) {{\textbf{$\{r_1==|i-j|\}$ {is SAT}}}}; \& \\
};
\tikzstyle{every path}=[line,distance=1cm]
\path[draw=red] (1) -- (4);
\path[draw=red] (4) -- (5);
\path (5.west) -- node [left] {If} (3.north);
\path[draw=red] (5.east) -- node [right] {Else} (6.north);
\path (3.south) -- (7);
\path[draw=red] (6.south) -- (7);
\path[draw=red] (7.west) -- node [left] {If} (8.north);
\path (7.east) -- node [right] {Else} (9.north);
\path[draw=red] (8.south) -- (10.west);
\path (9.south) -- (10.east);
\end{tikzpicture}
\vspace{-0.5cm}
\caption{{\scriptsize Le chemin d'une déviation non minimale:$\{i_0 \leq j_0,k_1=1 \land i_0!=j_0\}$}} 
\label{cfgAbsMinus4}
\end{tiny}
\end{minipage}
\begin{minipage}{2cm}
\end{minipage} \hfill
\vspace{-0.5cm}
\end{figure}

Le tableau ci-dessous récapitule le déroulement de {\tt LocFaults} pour le programme AbsMinus, avec au plus 2 conditions déviées à partir du contre-exemple suivant $\{i=0,j=1\}$. 
\begin{tiny}
\begin{tabular}{|c|c|c|c|}
  \hline
  \textit{Conditions déviées} & \textit{DCM}  & \textit{MCS} & Figure  \\
  \hline
  \multirow{2}{*}{$\emptyset$} & \multirow{2}{*}{/} & \multirow{2}{*}{$\{r_1=i_0-j_0:15\}$} & \multirow{2}{*}{fig. 3} \\
         &   &     &  \\   
  \hline
  $\{i_0 \leq j_0:8\}$ & Non & / & fig. 4 \\
  \hline
  \multirow{2}{*}{$\{k_1 = 1 \land i_0 != j_0:11\}$} & \multirow{2}{*}{Oui} & $\{k_0=0:7\}$, & \multirow{2}{*}{fig. 5} \\
         &  & $\{k_1=k_0+2:9\}$ &  \\  
  \hline
  $\{i_0 \leq j_0:8,$ & \multirow{2}{*}{Non}   & \multirow{2}{*}{/} & \multirow{2}{*}{fig. 6} \\  
  $ k_1 = 1 \land i_0 != j_0:11\}$  &  &  &  \\  
  \hline
\end{tabular}
\end{tiny}
Nous avons affiché les conditions déviées, si elles constituent une déviation minimale ou non, les MCSs calculés à partir du système construit : voir respectivement les colonnes 1, 2 et 3. La colonne 4 indique la figure qui illustre le chemin exploré pour chaque déviation. Sur la première et la troisième colonne, nous avons affiché en plus de l'instruction sa ligne dans le programme. Exemple, la première ligne dans le tableau montre qu'il y a un seul MCS trouvé ($\{r_1=i_0-j_0:15\}$) sur le chemin du contre-exemple.   

\section{Traitement des boucles} 
Dans le cadre du Bounded Model Checking (BMC) pour les programmes, le dépliage peut être appliqué au programme en entier comme il peut être appliqué aux boucles séparément~\cite{d2008survey}. Notre approche de localisation d'erreurs, {\tt LocFaults}~\cite{locfaults1}~\cite{locfaults2}, se place dans la deuxième démarche ; c'est-à-dire, nous utilisons une borne $b$ pour déplier les boucles en les remplaçant par des imbrications de conditionnelles de profondeur $b$. Considérons le programme Minimum (voir fig. 7) contenant une seule boucle, qui calcule le minimum dans un tableau d'entiers. L'effet sur le graphe de flot de contrôle du programme Minimum avant et après le dépliage est illustré sur les figures respectivement 7 et 8 : la boucle \textit{While} est dépliée 3 fois, tel que 3 est le nombre d'itérations nécessaires à la boucle pour calculer la valeur minimum dans un tableau de taille 4 dans le pire des cas.

{\tt LocFaults} prend en entrée le CFG du programme erroné, $CE$ un contre-exemple, $b_{dcm}$ : une borne sur le nombre de conditions déviées, $b_{mcs}$ : une borne sur la taille des MCSs calculés. Il permet d'explorer le CFG en profondeur en déviant au plus $b_{dcm}$ conditions par rapport au comportement du contre-exemple :
\begin{itemize}
\item[*] Il propage le contre-exemple jusqu'à la postcondition. Ensuite, il calcule les MCSs sur le CSP du chemin généré pour localiser les erreurs sur le chemin du contre-exemple. 
\item[*] Il cherche à énumérer les ensembles $\leq b_{dcm}$-DCM. Pour chaque DCM trouvée, il calcule les MCSs dans le chemin qui arrive à la dernière condition déviée et qui permet de prendre le chemin de la déviation.  
\end{itemize}

Parmi les erreurs les plus courantes associées aux boucles selon~\cite{debugloopswp}, le bug \textit{Off-by-one}, c'est-à-dire, des boucles qui s'itèrent une fois de trop ou de moins. Cela peut être dû à une mauvaise initialisation des variables de contrôle de la boucle, ou à une condition incorrecte de la boucle. Le programme \textit{Minimum} présente un cas de ce type d'erreur. Il est erroné à cause de sa boucle \textit{While}, l'instruction falsifiée se situe sur la condition de la boucle (ligne 9): la condition correcte doit être $(i<tab.length)$ ($tab.length$ est le nombre d'éléments du tableau $tab$). \`A partir du contre-exemple suivant : $\{tab[0]=3, tab[1]=2, tab[2]=1, tab[3]=0\}$, nous avons illustré sur la figure 8 le chemin fautif initial (voir le chemin coloré en rouge), ainsi que la déviation pour laquelle la postcondition est satisfaisable (la déviation ainsi que le chemin au-dessus de la condition déviée sont illustrés en vert).

%\begin{itemize}
%\item Variables non-initialisées correctement avant d'entrer dans une boucle ;
%\item Une boucle peut itérer zéro fois ;
%\item Une boucle peut être infinie.
%\end{itemize}

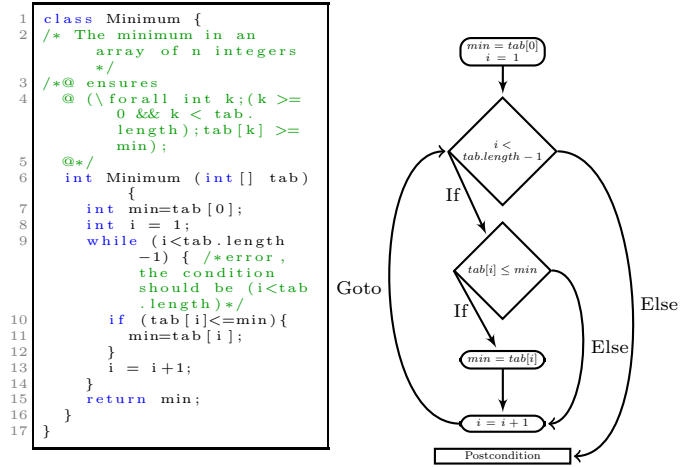
\begin{figure}[!ht]
\begin{minipage}{2cm}
\end{minipage} \hfill
\begin{minipage}[b]{.45\linewidth}
%Ci-dessous un programme simple avec une boucle \textit{While}. Le programme utilisé, nommé \textit{Minimum}, prend en entrée un tableau $tab$ des entiers, et devra retourner sa valeur minimum.
\begin{lstlisting}{}
class Minimum {
/* The minimum in an array of n integers*/
/*@ ensures 
  @ (\forall int k;(k >= 0 && k < tab.length);tab[k] >= min);
  @*/
  int Minimum (int[] tab) {
    int min=tab[0];
    int i = 1;
    while (i<tab.length-1) { /*error, the condition should be (i<tab.length)*/
      if (tab[i]<=min){
        min=tab[i];
      }
     	i = i+1;
    }
    return min;
  }
}
\end{lstlisting}

\end{minipage} \hfill
\begin{minipage}[b]{.52\linewidth}

\begin{scriptsize}
\begin{tikzpicture}[auto,node distance=2cm,/.style={font=\sffamily\footnotesize}]
\tikzstyle{decision} = [scale=0.6,diamond, draw=black, thick,text width=3em, text badly centered, inner sep=1pt]
\tikzstyle{block} = [scale=0.6,rectangle, draw=black, thick, 
text width=6em, text centered, rounded corners, minimum height=0.8em,text height=0.5em]
\tikzstyle{bloc} = [scale=0.6,rectangle, draw=black, thick, text centered, minimum height=0.8em,text height=0.5em,minimum width=3cm]
\tikzstyle{line} = [scale=0.6,draw, thick, -latex',shorten >=1pt];
\tikzstyle{cloud} = [scale=0.6,thick, ellipse,draw=black, minimum height=2em];
\matrix [column sep=0.5mm,row sep=2mm,ampersand replacement=\&]
{
\& \node [block] (5) {$min=tab[0]$  $i = 1$}; \& \\

\&  \&  \\

\& \node [decision,text width=6.3em] (6) {$i<tab.length-1$}; \& \\

\& \node [decision,text width=6.3em] (7) {$tab[i] \leq min$}; \& \\

\&  \&  \\

\& \node [block] (8) {$min=tab[i]$}; \&  \\

\&  \&  \\

\&  \&  \\

\& \node [block] (9) {$i=i+1$}; \& \\

\& \node [bloc] (18) {Postcondition}; \& \\
};
\tikzstyle{every path}=[line,distance=0.5cm]
%\path (1) -- (4);
\path (5) -- (6);
\path (6.west) -- node [left] {If} (7);
\path (7.west) -- node [left] {If} (8);
\path (8) -- (9.north);
\draw[->] (7.east) .. controls +(right:10mm) and +(right:10mm) .. node [right] {Else} (9.east);
\draw[->] (6.east) .. controls +(right:20mm) and +(right:20mm) .. node [right] {Else} (18.east);
\draw[->] (9.west) .. controls +(left:20mm) and +(left:30mm) .. node [left] {Goto} (6);
\end{tikzpicture}
\end{scriptsize}
\end{minipage}

\label{MinimumCFG}
\vspace{-0.4cm}
\caption{Le programme Minimum et son CFG normal (non déplié). La postcondition est $\{\forall\text{ int k};(k\geq0 \land k<tab.length);tab[k] \geq min\}$}
\end{figure}

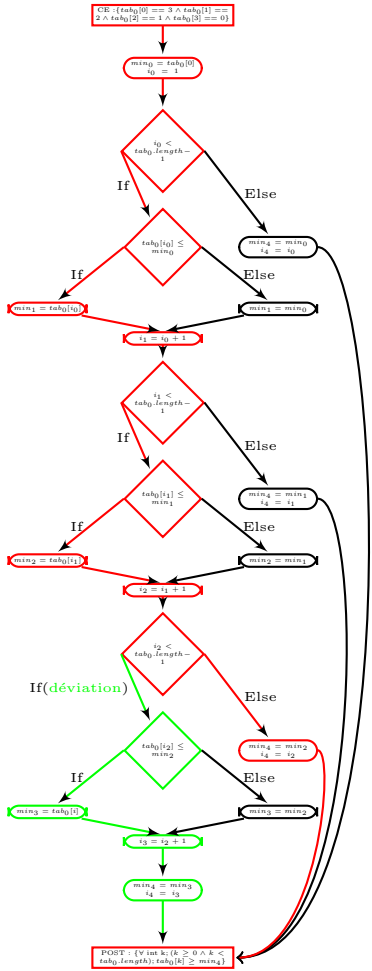
\begin{figure}[!ht]
\center
\begin{tiny}
\begin{tikzpicture}[scale=0.5,auto,node distance=2cm,/.style={font=\sffamily\footnotesize}]
\tikzstyle{decision} = [scale=0.5,diamond, draw=black, thick,text width=4em, text badly centered, inner sep=1pt]
\tikzstyle{block} = [scale=0.5,rectangle, draw=black, thick, 
text width=8em, text centered, rounded corners, minimum height=0.8em,text height=0.5em];
\tikzstyle{bloc} = [scale=0.5,rectangle, draw=black, thick, 
text width=15em, text centered, minimum height=0.8em,text height=0.5em];
\tikzstyle{line} = [scale=0.5,draw, thick, -latex',shorten >=1pt];
\tikzstyle{cloud} = [scale=0.5,thick, ellipse,draw=black, minimum height=2em];
\matrix [column sep=0.5mm,row sep=2mm,ampersand replacement=\&]
{

\& \node [bloc,draw=red] (22) {CE:$\{tab_0[0]==3 \land tab_0[1]==2 \land tab_0[2]==1 \land tab_0[3]==0\}$}; \& \\

\&  \&  \\

\& \node [block,draw=red] (5) {$min_0=tab_0[0]$  $i_0 = 1$}; \& \\

\&  \&  \\

\& \node [decision,text width=6.3em,draw=red] (6) {$i_0$ $<$ $tab_{0}.length-1$}; \& \\

\& \node [decision,text width=6.3em,draw=red] (7) {$tab_{0}[i_0]$ $\leq$ $min_0$}; \& \node [block] (23) {$min_4=min_0$  $i_4=i_0$};  \\

\node [block,draw=red] (8) {$min_{1}$ $=$ $tab_{0}[i_0]$}; \&  \& \node [block] (19) {$min_{1}$ $=$ $min_{0}$}; \\

\& \node [block,draw=red] (9) {$i_{1}$ $=$ $i_{0}+1$}; \& \\

\& \node [decision,text width=6.3em,draw=red] (10) {$i_1$ $<$ $tab_{0}.length-1$}; \& \\

\& \node [decision,text width=6.3em,draw=red] (11) {$tab_{0}[i_1]$ $\leq$ $min_1$}; \& \node [block] (24) {$min_4=min_1$  $i_4=i_1$}; \\

\node [block,draw=red] (12) {$min_2$ $=$ $tab_{0}[i_1]$}; \&  \& \node [block] (20) {$min_{2}$ $=$ $min_{1}$};  \\

\& \node [block,draw=red] (13) {$i_2$ $=$ $i_1+1$}; \& \\

\& \node [decision,text width=6.3em,draw=red] (14) {$i_2$ $<$ $tab_{0}.length-1$}; \& \\

\& \node [decision,text width=6.3em,draw=green] (15) {$tab_{0}[i_2]$ $\leq$ $min_2$}; \& \node [block,draw=red] (25) {$min_4=min_2$  $i_4=i_2$}; \\

\node [block,draw=green] (16) {$min_3$ $=$ $tab_{0}[i]$}; \&  \& \node [block] (21) {$min_{3}$ $=$ $min_{2}$};  \\

\& \node [block,draw=green] (17) {$i_3$ $=$ $i_2+1$}; \& \\

\&  \& \\

\& \node [block,draw=green] (26) {$min_4=min_3$  $i_4=i_3$}; \& \\

\&  \& \\

\&  \& \\ 

\& \node [bloc,draw=red] (18) {POST : $\{\forall\text{ int k};(k\geq0 \land k<tab_{0}.length);tab_{0}[k] \geq min_4\}$}; \& \\
};
\tikzstyle{every path}=[line,distance=1cm]
%\path (1) -- (4);
\path[draw=red] (22) -- (5);
\path[draw=red] (5) -- (6);
\path[draw=red] (6.west) -- node [left] {If} (7);
\path (6.east) -- node [right] {Else} (23);
\path[draw=red] (7.west) -- node [left] {If} (8);
\path (7.east) -- node [right] {Else} (19);
\path[draw=red] (8) -- (9.north);
\path (19) -- (9.north);
\path[draw=red] (9) -- (10);
\path[draw=red] (10.west) -- node [left] {If} (11);
\path (10.east) -- node [right] {Else} (24);
\path[draw=red] (11.west) -- node [left] {If} (12);
\path (11.east) -- node [right] {Else} (20);
\path[draw=red] (12) -- (13.north);
\path (20) -- (13.north);
\path[draw=red] (13) -- (14);
\path[draw=green] (14.west) -- node [left] {If(\textcolor{green}{déviation})} (15);
\path[draw=red] (14.east) -- node [right] {Else} (25);
\path[draw=green] (15.west) -- node [left] {If} (16);
\path (15.east) -- node [right] {Else} (21);
\path[draw=green] (16) -- (17.north);
\path (21) -- (17.north);
\path[draw=green] (17) -- (26);
\path[draw=green] (26) -- (18);
\draw[->] (23.east) .. controls +(right:35mm) and +(right:95mm) ..  (18.east);
\draw[->] (24.east) .. controls +(right:25mm) and +(right:70mm) ..  (18.east);
\draw[->,draw=red] (25.east) .. controls +(right:10mm) and +(right:45mm) ..  (18.east);
\end{tikzpicture}
\end{tiny}
\label{MinimumCFGDeplie}
\vspace{-0.4cm}
\caption{{\scriptsize Figure montrant le CFG en forme DSA du programme \textit{Minimum} en dépliant sa boucle $3$ fois, avec le chemin d'un contre-exemple (illustré en rouge) et une déviation satisfaisant sa postcondition (illustrée en vert).}}
\vspace{-0.6cm}
\end{figure}

Nous affichons dans le tableau ci-dessous les chemins erronés générés (la colonne $PATH$) ainsi que les MCSs calculés (la colonne $MCSs$) pour au plus $1$ condition déviée par rapport au comportement du contre-exemple. La première ligne correspond au chemin du contre-exemple ; la deuxième correspond au chemin obtenu en déviant la condition $\{i_2 \leq tab_0.length-1\}$.

\begin{tiny}
\begin{tabular}{|c|c|}
\hline
 $PATH$ & MCSs \\ 
\hline
 $\{CE:[tab_0[0]=3 \land tab_0[1]=2 \land tab_0[2]=1$  & \multirow{5}{*}{$\{min_2=tab_0[i_1]\}$}  \\
 $\land  tab_0[3]==0]$, $min_0=tab_0[0]$, $i_0=1$, &  \\
 $min_1=tab_0[i_0]$,$i_1=i_0+1$,$min_2=tab_0[i_1]$, &  \\
 $i_2=i_1+1$,$min_3=min_2$, $i_3=i_2$,  &   \\ 
 $POST:[(tab[0] \geq min_3) \land  (tab[1] \geq min_3)$  &  \\
 $\land (tab[2] \geq min_3) \land (tab[3] \geq min_3)]\}$ &  \\
\hline
 $\{CE:[tab_0[0]=3 \land tab_0[1]=2 \land tab_0[2]=1$  & \multirow{2}{*}{$\{i_0=1\}$,}  \\
 $\land  tab_0[3]==0]$, $min_0=tab_0[0]$, $i_0=1$, & \multirow{2}{*}{$\{i_1=i_0+1\}$,} \\
 $min_1=tab_0[i_0]$,$i_1=i_0+1$,$min_2=tab_0[i_1]$,  &  \multirow{2}{*}{$\{i_2=i_1+1\}$} \\
 $i_2=i_1+1$,$[\neg (i_2 \leq tab_0.length-1)]$  &  \\
\hline
\end{tabular}
\end{tiny}

{\tt LocFaults} a permis d'identifier un seul MCS sur le chemin du contre-exemple qui contient la contrainte $min_2=tab_0[i_1]$, l'instruction de la ligne $11$ dans la deuxième itération de la boucle dépliée. Avec une condition déviée, l'algorithme suspecte la troisième condition de la boucle dépliée, $i_2$ $<$ $tab_{0}.length-1$ ; en d'autres termes, il faut une nouvelle itération pour satisfaire la postcondition. 

Cet exemple montre un cas d'un programme avec une boucle erronée : l'erreur est sur le critère d'arrêt, elle ne permet pas en effet au programme d'itérer jusqu'au dernier élément du tableau en entrée. {\tt LocFaults} avec son mécanisme de déviation arrive à supporter ce type d'erreur avec précision. Il fournit à l'utilisateur non seulement les instructions suspectes dans la boucle non dépliée du programme original, mais aussi des informations sur les itérations où elles se situent concrètement en dépliant la boucle. Ces informations pourraient être très utiles pour le programmeur pour mieux comprendre les erreurs dans la boucle.

\section{Algorithme amélioré}
Notre but consiste à trouver les DCMs de taille inférieure à une borne $k$ ; en d'autres termes, on cherche à donner une solution au problème posé ci-dessus ($\leq k$-DCM). Pour cela, notre algorithme (nommé {\tt LocFaults}) parcourt en profondeur le CFG et génère les chemins où au plus $k$ conditions sont déviées par rapport au comportement du contre-exemple.  

Pour améliorer l'efficacité, notre solution heuristique procède de façon incrémentale. Elle dévie successivement de $0$ à $k$ conditions et elle recherche les MCSs pour les chemins correspondants. Toutefois, si à l'étape $k$ {\tt LocFaults} a dévié une condition $c_i$ et que cela a corrigé le programme, elle n'explorera pas à l'étape $k'$ avec $k'>k$ les chemins qui impliquent une déviation de la condition $c_i$. Pour cela, nous ajoutons la cardinalité de la déviation minimale trouvée ($k$) comme information sur le n\oe{}ud de $c_i$.

Nous allons sur un exemple illustrer le déroulement de notre approche, voir le graphe sur la figure 9. Chaque cercle dans le graphe représente un n\oe{}ud conditionnel visité par l'algorithme. L'exemple ne montre pas les blocs d'affectations, car nous voulons illustrer uniquement comment nous trouverons les déviations de correction minimales d'une taille bornée de la manière citée ci-dessus. Un arc reliant une condition $c_1$ à une autre $c_2$ illustre que $c_2$ est atteinte par l'algorithme. Il y a deux façons, par rapport au comportement du contre-exemple, par lesquelles {\tt LocFaults} arrive à la condition $c_2$:
\begin{enumerate}
\item en suivant la branche normale induite par la condition $c_1$ ;
\item en suivant la branche opposée.
\end{enumerate}
La valeur de l'étiquette des arcs pour le cas (1) (resp. (2)) est "\textit{next}" (resp. "\textit{devie}"). 
 
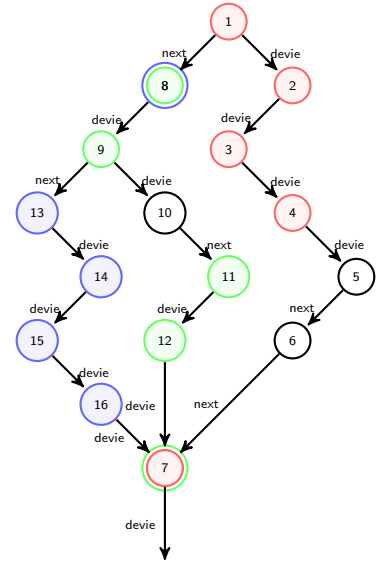
\begin{figure}[!ht]
\centering
\begin{tikzpicture}[->,>=stealth',shorten >=1pt,auto,node distance=1.2cm,thick,main node/.style={circle,draw,font=\sffamily\tiny}]
  \node[main node,draw=red!60, fill=red!5] (1) {1};
  \node[main node,draw=green!60, fill=green!5] (2) [below left of=1] {8};
  \node[main node,draw=blue!60, minimum size=6mm] (18) [below left of=1] {8};
  \node[main node,draw=green!60, fill=green!5] (3) [below left of=2] {9};
  \node[main node] (4) [below right of=3] {10};
  \node[main node,draw=green!60, fill=green!5] (6) [below right of=4] {11};
  \node[main node,draw=green!60, fill=green!5] (7) [below left of=6] {12};
  \node[main node,draw=red!60, fill=red!5] (9) [below right of=1] {2};
  \node[main node,draw=red!60, fill=red!5] (10) [below left of=9] {3};
  \node[main node,draw=red!60, fill=red!5] (11) [below right of=10] {4};
  \node[main node] (12) [below right of=11] {5};
  \node[main node] (13) [below left of=12] {6};
  \node[main node,draw=blue!60, fill=blue!5] (14) [below left of=3] {13};
  \node[main node,draw=blue!60, fill=blue!5] (15) [below right of=14] {14};
  \node[main node,draw=blue!60, fill=blue!5] (16) [below left of=15] {15};
  \node[main node,draw=blue!60, fill=blue!5] (17) [below right of=16] {16};  
  \node[main node,draw=red!60, fill=red!5] (8) [below right of=17] {7};
  \node[main node,draw=green!60, minimum size=6mm] (19) [below right of=17] {7};
  \node[main node,draw=red!60, fill=red!5] (20) [below right of=17] {7};
  \node[node distance=1.5cm,rectangle] (21) [below of=8] {{\scriptsize le chemin $<1,2,3,4,5,6,7,...,POST>$ est correct}};
  \node[node distance=0.4cm,rectangle] (22) [below of=21] {{\scriptsize le chemin $<1,8,9,10,11,12,7,...,POST>$ est correct}};
  \path[every node/.style={font=\sffamily\tiny}]
    (1) edge node [left] {next} (2)
    (18) edge node [left] {devie} (3) 
    (3) edge node [right] {devie} (4)
    (4) edge node [right] {next} (6)
    (6) edge node [left] {devie} (7)
    (7) edge node [left] {devie} (8)
    (1) edge node [right] {devie} (9)
    (9) edge node [left] {devie} (10)
    (10) edge node [right] {devie} (11)  
    (11) edge node [right] {devie} (12)
    (12) edge node [left] {next} (13)
    (13) edge node [left] {next} (8)         
    (3) edge node [left] {next} (14)
    (14) edge node [right] {devie} (15)
    (15) edge node [left] {devie} (16)
    (16) edge node [right] {devie} (17)
    (17) edge node [left] {devie} (8)     
    (8) edge node [left] {devie} (21);
\end{tikzpicture}
\label{cfgExempleAbstrait}
\caption{{\footnotesize Figure illustrant l'exécution de notre algorithme sur un exemple pour lequel deux déviations minimales sont détectées: $\{1,2,3,4,7\}$ et $\{8,9,11,12,7\}$, et une abandonnée: $\{8,13,14,15,16,7\}$. Sachant que la déviation de la condition "7" a permis de corriger le programme pour le chemin $<1,2,3,4,5,6>$, ainsi que pour le chemin $<1,8,9,10,11,12,7>$.}}
\end{figure}

\begin{itemize}
\item \`A l'étape $k=5$, notre algorithme a identifié deux déviations minimales de taille égale à $5$:
\begin{enumerate}
\item $D_1=\{1,2,3,4,7\}$, le n\oe{}ud "$7$" est marqué par la valeur $5$ ;
\item $D_2=\{8,9,11,12,7\}$, elle a été autorisée, car la valeur de la marque du n\oe{}ud "$7$" est égale à la cardinalité de $D_2$.  
\end{enumerate}

\item \`A l'étape $k=6$, l'algorithme a suspendu la déviation suivante $D_3=\{8,13,14,15,16,7\}$, car la cardinalité de $D_3$ est supérieure strictement à la valeur de l'étiquette du n\oe{}ud "$7$". 
\end{itemize}

\section{Expérience pratique} 
Pour évaluer la scalabilité de notre méthode, nous avons comparé ses performances avec celles de {\tt BugAssist}\footnote{L'outil BugAssist est disponible à l'adresse : \url{http://bugassist.mpi-sws.org/}} sur deux ensembles de benchmarks\footnote{Le code source de l'ensemble de programmes est disponible à l'adresse : \url{http://www.i3s.unice.fr/~bekkouch/Benchs_Mohammed.html}}.
\begin{itemize}
\item[*] Le premier benchmark est illustratif, il contient un ensemble de programmes sans boucles ;
\item[*] Le deuxième benchmark inclut 19, 49 et 91 variations pour respectivement les programmes BubbleSort, Sum et SquareRoot. Ces programmes contiennent des boucles pour étudier le passage à l'échelle de notre approche par rapport à {\tt BugAssist}. Pour augmenter la complexité d'un programme, nous augmentons le nombre d'itérations dans les boucles à l'exécution de chaque outil ; nous utilisons la même borne de dépliage des boucles pour {\tt LocFaults} et {\tt BugAssist}. 
\end{itemize}

Pour générer le CFG et le contre-exemple, nous utilisons l'outil CPBPV~\cite{cpbpv2010} (Constraint-Programming Framework for Bounded Program Verification). {\tt LocFaults} et {\tt BugAssist} travaillent respectivement sur des programmes Java et C. Pour que la comparaison soit juste, nous avons construit pour chaque programme deux versions équivalentes :
\begin{itemize}
\item[*] une version en Java annotée par une spécification JML ;
\item[*] une version en ANSI-C annotée par la même spécification mais en ACSL.
\end{itemize}
Les deux versions ont les mêmes numéros de lignes d'instructions, notamment des erreurs. La précondition spécifie le contre-exemple employé pour le programme.

Pour calculer les MCSs, nous avons utilisé les solveurs IBM ILOG MIP\footnote{ Disponible à l'adresse http://www-01.ibm.com/software/commerce/optimization/cplex-optimizer/} et CP\footnote{Disponible à l'adresse http://www-01.ibm.com/software/commerce/optimization/cplex-cp-optimizer/} de CPLEX. Nous avons adapté et implémenté l'algorithme de Liffiton et Sakallah~\cite{liffiton2008}, voir alg.~\ref{liffiton2008}. Cette implémentation prend en entrée l'ensemble de contraintes infaisable qui correspond au chemin identifié ($C$), et $b_{mcs}$: la borne sur la taille des MCSs calculés. Chaque contrainte $c_i$ dans le système construit $C$ est augmentée par un indicateur $y_i$ pour donner $y_i \rightarrow c_i$ dans le nouveau système de contraintes $C'$. Affecter à $y_i$ la valeur $Vrai$ implique la contrainte $c_i$ ; en revanche, affecter à $y_i$ la valeur $Faux$ implique la suppression de la contrainte $c_i$. Un MCS est obtenu en cherchant une affectation qui satisfait le système de contraintes avec un ensemble minimal d'indicateurs de contraintes affectés avec $Faux$. Pour limiter le nombre de variables indicateurs de contraintes qui peuvent être assignées à Faux, on utilise la contrainte $AtMost({\neg y_1,\neg y_2,...,\neg y_n},k)$ (voir la ligne $5$), le système créé est noté dans l'algorithme $C'_{k}$ (ligne $5$). Chaque itération de la boucle \textsc{While} (lignes $6-19$) permet de trouver tous les MCSs de taille $k$, $k$ est incrémenté de 1 après chaque itération. Après chaque MCS trouvé (lignes $8-13$), une contrainte de blocage est ajoutée à $C'_k$ et $C'$ pour empêcher de trouver ce nouveau MCS dans les prochaines itérations (lignes $15-16$). La première boucle (lignes $4-19$) s'itère jusqu'à ce que tous les MCSs de $C$ soient générés ($C'$ devient infaisable) ; elle peut s'arrêter aussi si les MCSs de taille inférieure ou égale $b_{mcs}$ sont obtenus ($k > b_{mcs}$).

\begin{algorithm}[h]
\begin{tiny} 
\label{MCS}
{\tt Fonction MCS}($C$,$b_{mcs}$)\\
\Entree{$C$: Ensemble de contraintes infaisable, $b_{mcs}$: Entier}
\Sortie{$MCS$: Liste de MCSs de $C$ de cardinalité inférieure à $b_{mcs}$}
\Deb{
$C'$ $\leftarrow$ \textsc{AddYVars}($C$); 
$MCS$ $\leftarrow$ $\emptyset$; 
$k$ $\leftarrow$ $1$; \\ 
\Tq{\textsc{SAT}($C'$) $\land$ $k \leq b_{mcs}$}{
$C'_{k}$ $\leftarrow$ $C'$ $\land$ \textsc{AtMost}($\{\neg y_{1},\neg y_{2},...,\neg y_{n}\}$,$k$)\\
\Tq{SAT($C'_{k}$)}{
$newMCS$ $\leftarrow$ $\emptyset$\\
\PourCh{indicateur $y_{i}$}{
\% $y_{i}$ est l'indicateur de la contrainte $c_{i} \in C$, et $val(y_{i})$ la valeur de $y_{i}$ dans la solution calculée de $C'_{k}$.\\
\Si{$val(y_{i})=0$}{
$newMCS$ $\leftarrow$ $newMCS$ $\cup$ $\{c_{i}\}$.\\
}
}
$MCS.add(newMCS)$.\\ 
$C'_{k}$ $\leftarrow$ $C'_{k}$ $\land$ \textsc{BlockingClause}($newMCS$) % {\scriptsize On ajoute la contrainte de blocage de $newMCS$ à l'ensemble de contraintes $C'_{k}$}
\\
$C'$ $\leftarrow$ $C'$ $\land$ \textsc{BlockingClause}($newMCS$) % {\scriptsize On ajoute la contrainte de blocage de $newMCS$ à l'ensemble de contraintes $C'$}\\
}
$k$ $\leftarrow$ $k$ + $1$\\
}
\Retour $MCS$
}
\end{tiny}
\label{liffiton2008}
\caption{Algorithme de Liffiton et Sakallah}
\end{algorithm}

{\tt BugAssist} utilise l'outil CBMC~\cite{cbmc2004} pour générer la trace erronée et les données d'entrée. Pour le solveur Max-SAT, nous avons utilisé MSUnCore2~\cite{msuncore2}.

Les expérimentations ont été effectuées avec un processeur Intel Core i7-3720QM 2.60 GHz avec 8 GO de RAM.

\subsection{Le benchmark sans boucles}
%Tritype est un benchmark standard sans boucle avec plusieurs instructions conditionnelles. Le programme prend trois entiers positifs comme entrées $(i,j,k)$ représentant les cotés d'un triangle, et retourne 2 si les entrées correspondent à un triangle isocèle, la valeur 3 si elles correspondent à un triangle équilatéral, la valeur 4 si elles correspondent à un autre triangle, la valeur 4 si les entrée correspondent à un triangle non-valide.   

Cette partie sert à illustrer l'amélioration apportée à {\tt LocFaults} pour réduire le nombre d'ensembles suspects fournis à l'utilisateur : à une étape donnée de l'algorithme, le n\oe{}ud dans le CFG du programme qui permet de détecter une DCM sera marqué par le cardinal de cette dernière ; ainsi aux prochaines étapes, l'algorithme n'autorisera pas le balayage d'une liste d'adjacence de ce n\oe{}ud.

Nos résultats\footnote{Le tableau qui donne les MCSs calculés par {\tt LocFaults} pour les programmes sans boucles est disponible à l'adresse \url{http://www.i3s.unice.fr/~bekkouch/Bench_Mohammed.html#rsb}} montrent que {\tt LocFaults} rate les erreurs uniquement pour TritypeKO6. Or, {\tt BugAssist} rate l'erreur pour AbsMinusKO2, AbsMinusKO3, AbsMinusV2KO2, TritypeKO, TriPerimetreKO, TriMultPerimetreKO et une des deux erreurs dans TritypeKO5. Les temps\footnote{Les tableaux qui donnent les temps de {\tt LocFaults} et {\tt BugAssist} pour les programmes sans boucles sont disponibles à l'adresse \url{http://www.i3s.unice.fr/~bekkouch/Bench_Mohammed.html#rsba}} de notre outil sont meilleurs par rapport à {\tt BugAssist} pour les programmes avec calcul numérique ; ils sont proches pour le reste des programmes.

Prenons trois exemples parmi ces programmes au hasard. Et considérons l'implémentation de deux versions de notre algorithme, sans et avec marquage des n\oe{}uds nommées respectivement {\tt LocFaultsV1} et {\tt LocFaultsV2}.

\begin{itemize}
\item Les tables~\ref{MCSeskCF} et~\ref{time1} montrent respectivement les ensembles suspects et les temps de {\tt LocFaultsV1} ;
\item Les tables~\ref{MCSeskCFmarke} et~\ref{time1marke} montrent respectivement les ensembles suspects et les temps de {\tt LocFaultsV2}.
\end{itemize}
Dans les tables ~\ref{MCSeskCF} et~\ref{MCSeskCFmarke}, nous avons affiché la liste des MCSs et DCMs calculés. Le numéro de la ligne correspondant à la condition est souligné. Les tables~\ref{time1} et~\ref{time1marke} donnent les temps de calcul : $P$ est le temps de prétraitement qui inclut la traduction du programme Java en un arbre syntaxique abstrait avec l'outil JDT (Eclipse Java devlopment tools), ainsi que la construction du CFG ; $L$ est le temps de l'exploration du CFG et de calcul des MCSs.     

{\tt LocFaultsV2} a permis de réduire considérablement les déviations générées ainsi que les temps sommant l'exploration du CFG et le
calcul des MCSs de {\tt LocFaultsV1}, et cela sans perdre l'erreur ; les localisations fournies par {\tt LocFaultsV2} sont plus pertinentes. Les lignes éliminées de la table~\ref{MCSeskCFmarke} sont colorées en bleu dans la table~\ref{MCSeskCF}. Les temps améliorés sont affichés en gras dans la table~\ref{time1marke}. Par exemple, pour le programme TritypeKO2, à l'étape $1$ de l'algorithme, {\tt LocFaultsV2} marque le n\oe{}ud de la condition $26$, $35$ et $53$ (à partir du contre-exemple, le programme devient correct en déviant chacune de ces trois conditions). Cela permet, à l'étape $2$, d'annuler les déviations suivantes: $\{\uline{26},\uline{29}\}$, $\{\uline{26},\uline{35}\}$, $\{\uline{29},\uline{35}\}$, $\{\uline{32},\uline{35}\}$. Toujours à l'étape $2$, {\tt LocFaultsV2} détecte deux déviations minimales en plus: $\{\uline{29},\uline{57}\}$, $\{\uline{32},\uline{44}\}$, les n\oe{}uds $57$ et $44$ vont donc être marqués (la valeur de la marque est $2$). \`A l'étape $3$, aucune déviation n'est sélectionnée ; à titre d'exemple, $\{\uline{29},\uline{32},\uline{44}\}$ n'est pas considérée parce que son cardinal est supérieur strictement à la valeur de la marque du n\oe{}ud $44$.

\begin{table*}
\begin{center}
\begin{tiny}
\begin{tabular}{|c|c|c|c|c|c|c|}
\hline
\multirow{2}{*}{Programme} & \multirow{2}{*}{Contre-exemple} & \multirow{2}{*}{Erreurs} & \multicolumn{4}{|c|}{LocFaults} \\ 
\cline{4-7}
  &   &   & $= 0$ & $\leq 1$ & $\leq 2$ & $\leq 3$  \\   
\hline
\multirow{15}{*}{TritypeKO2}  &  \multirow{15}{*}{$\{i=2,j=2,k=4\}$}  &  \multirow{15}{*}{$53$}  &  \multirow{15}{*}{$\{54\}$}  &  $\{54\}$   & $\{54\}$   &  $\{54\}$   \\ 
\cline{6-7}
\cline{5-5}
&    &    &     & $\{\uline{21}\}$   & $\{\uline{21}\}$  &  $\{\uline{21}\}$   \\
\cline{6-7}
\cline{5-5}
&    &    &     & $\{\uline{26}\}$ & $\{\uline{26}\}$  &  $\{\uline{26}\}$  \\
\cline{5-7}
&    &    &     &  $\{\uline{35}\}$,$\{27\}$,$\{25\}$  & $\{\uline{35}\}$,$\{27\}$,$\{25\}$    & $\{\uline{35}\}$,$\{27\}$,$\{25\}$  \\
\cline{6-7}
\cline{5-5}
&    &    &     &  \multirow{11}{*}{$\{\uline{\textcolor{red}{53}}\}$,$\{25\}$,$\{27\}$}  &  $\{\uline{\textcolor{red}{53}}\}$,$\{25\}$,$\{27\}$     & $\{\uline{\textcolor{red}{53}}\}$,$\{25\}$,$\{27\}$       \\
\cline{6-7}
&    &    &     &     & \color{blue!90}$\{\uline{26},\uline{29}\}$  & \color{blue!90}$\{\uline{26},\uline{29}\}$     \\
\cline{6-7}
&    &    &     &     & \color{blue!90}$\{\uline{26},\uline{35}\}$,$\{25\}$ & \color{blue!90}$\{\uline{26},\uline{35}\}$,$\{25\}$    \\
\cline{6-7}
&    &    &     &     & \color{blue!90}$\{\uline{29},\uline{35}\}$,$\{30\}$,$\{25\}$,$\{27\}$  & \color{blue!90}$\{\uline{29},\uline{35}\}$,$\{30\}$,$\{25\}$,$\{27\}$     \\
\cline{6-7}
&    &    &     &     & $\{\uline{29},\uline{57}\}$,$\{30\}$,$\{27\}$,$\{25\}$   & $\{\uline{29},\uline{57}\}$,$\{30\}$,$\{27\}$,$\{25\}$   \\ 
\cline{6-7}
&    &    &     &     & \color{blue!90}$\{\uline{32},\uline{35}\}$,$\{33\}$,$\{25\}$,$\{27\}$   & \color{blue!90}$\{\uline{32},\uline{35}\}$,$\{33\}$,$\{25\}$,$\{27\}$   \\ 
\cline{6-7}
&    &    &     &     &  \multirow{6}{*}{$\{\uline{32},\uline{44}\}$,$\{33\}$,$\{25\}$,$\{27\}$}   &  $\{\uline{32},\uline{44}\}$,$\{33\}$,$\{25\}$,$\{27\}$   \\    
\cline{7-7}
&    &    &     &     &   & \color{blue!90}$\{\uline{26},\uline{29},\uline{35}\}$,$\{30\}$,$\{25\}$ \\ 
\cline{7-7}
&    &    &     &     &   & \color{blue!90}$\{\uline{26},\uline{32},\uline{35}\}$,$\{33\}$,$\{25\}$ \\
\cline{7-7}
&    &    &     &     &   & \color{blue!90}$\{\uline{26},\uline{32},\uline{57}\}$,$\{25\}$,$\{33\}$ \\
\cline{7-7}
\hhline{~~~~~~|6-6}
&    &    &     &     &   & \color{blue!90}$\{\uline{29},\uline{32},\uline{35}\}$,$\{33\}$,$\{25\}$,$\{27\}$,$\{30\}$ \\
\cline{7-7}
&    &    &     &     &   &  \color{blue!90}$\{\uline{29},\uline{32},\uline{44}\}$,$\{33\}$,$\{25\}$,$\{27\}$,$\{30\}$  \\
\hline
\multirow{14}{*}{TritypeKO4}  &  \multirow{14}{*}{$\{i=2,j=3,k=3\}$}  &  \multirow{14}{*}{$45$}  &  \multirow{14}{*}{$\{46\}$}  &  $\{46\}$   & $\{46\}$   &  $\{46\}$     \\
\cline{5-7}
&    &    &     &   \multirow{13}{*}{$\{\uline{\textcolor{red}{45}}\}$,$\{33\}$,$\{25\}$}  & $\{\uline{\textcolor{red}{45}}\}$,$\{33\}$,$\{25\}$   & $\{\uline{\textcolor{red}{45}}\}$,$\{33\}$,$\{25\}$  \\
\cline{6-7}
&    &    &     &     & $\{\uline{26},\uline{32}\}$   &  $\{\uline{26},\uline{32}\}$  \\
\cline{6-7}
&    &    &     &     & $\{\uline{29},\uline{32}\}$  & $\{\uline{29},\uline{32}\}$   \\
\cline{6-7}
&    &    &     &     &  \color{blue!90} $\{\uline{45},\uline{49}\}$,$\{33\}$,$\{25\}$  & \color{blue!90} $\{\uline{45},\uline{49}\}$,$\{33\}$,$\{25\}$   \\
\cline{6-7}
&    &    &     &     &  \color{blue!90}  & \color{blue!90} $\{\uline{45},\uline{53}\}$,$\{33\}$,$\{25\}$   \\
\cline{7-7}
&    &    &     &     &  \color{blue!90}  & \color{blue!90} $\{\uline{26},\uline{45},\uline{49}\}$,$\{33\}$,$\{25\}$,$\{27\}$   \\
\cline{7-7}
&    &    &     &     &  \color{blue!90}  & \color{blue!90} $\{\uline{26},\uline{45},\uline{53}\}$,$\{33\}$,$\{25\}$,$\{27\}$   \\
\cline{7-7}
&    &    &     &     &  \color{blue!90}  & \color{blue!90} $\{\uline{26},\uline{45},\uline{57}\}$,$\{33\}$,$\{25\}$,$\{27\}$   \\
\cline{7-7}
\hhline{~~~~~~|6-6}
&    &    &     &     &  \color{blue!90}  & \color{blue!90} $\{\uline{29},\uline{32},\uline{49}\}$,$\{30\}$,$\{25\}$   \\
\cline{7-7}
\hhline{~~~~~~|6-6}
&    &    &     &     &  \color{blue!90} $\{\uline{45},\uline{53}\}$,$\{33\}$,$\{25\}$ &  \color{blue!90} $\{\uline{29},\uline{45},\uline{49}\}$,$\{33\}$,$\{25\}$,$\{30\}$   \\
\cline{7-7}
\hhline{~~~~~~|6-6}
&    &    &     &     &  \color{blue!90}  & \color{blue!90} $\{\uline{29},\uline{45},\uline{53}\}$,$\{33\}$,$\{25\}$,$\{30\}$   \\
\cline{7-7}
\hhline{~~~~~~|6-6}
&    &    &     &     &  \color{blue!90}  & \color{blue!90} $\{\uline{29},\uline{45},\uline{57}\}$,$\{33\}$,$\{25\}$,$\{30\}$   \\
\cline{7-7}
&    &    &     &     &  \color{blue!90}  & $\{\uline{32},\uline{35},\uline{49}\}$,$\{25\}$    \\
\cline{7-7}
&    &    &     &     &  \color{blue!90}  & $\{\uline{32},\uline{35},\uline{53}\}$,$\{25\}$      \\
\cline{7-7}
&    &    &     &     &  \color{blue!90}  & $\{\uline{32},\uline{35},\uline{57}\}$,$\{25\}$     \\
\hline
\multirow{14}{*}{TriPerimetreKO3}  &  \multirow{14}{*}{$\{i=2,j=1,k=2\}$}  &  \multirow{14}{*}{$57$}   &  \multirow{14}{*}{$\{58\}$}  &  $\{58\}$  & $\{58\}$  &  $\{58\}$   \\
\cline{5-7}
&    &    &     & $\{\uline{22}\}$  &  $\{\uline{22}\}$   &  $\{\uline{22}\}$    \\
\cline{5-7}
&    &    &     &  $\{\uline{31}\}$  & $\{\uline{31}\}$  &  $\{\uline{31}\}$   \\
\cline{5-7}
&    &    &     &  $\{\uline{37}\}$,$\{32\}$,$\{27\}$   &  $\{\uline{37}\}$,$\{32\}$,$\{27\}$   &   $\{\uline{37}\}$,$\{32\}$,$\{27\}$  \\
\cline{5-7}
&    &    &     &  \multirow{10}{*}{$\{\textcolor{red}{\uline{57}}\}$,$\{32\}$,$\{27\}$}  &  $\{\textcolor{red}{\uline{57}}\}$,$\{32\}$,$\{27\}$  & $\{\textcolor{red}{\uline{57}}\}$,$\{32\}$,$\{27\}$   \\
\cline{6-7}
&    &    &     &     & \color{blue!90} $\{\uline{28},\uline{37}\}$,$\{32\}$,$\{27\}$,$\{29\}$   &  \color{blue!90} $\{\uline{28},\uline{37}\}$,$\{32\}$,$\{27\}$,$\{29\}$ \\ 
\cline{6-7}
&    &    &     &     & $\{\uline{28},\uline{61}\}$,$\{32\}$,$\{27\}$,$\{29\}$   &  $\{\uline{28},\uline{61}\}$,$\{32\}$,$\{27\}$,$\{29\}$    \\  
\cline{6-7}
&    &    &     &     & \color{blue!90} $\{\uline{31},\uline{37}\}$,$\{27\}$   & \color{blue!90} $\{\uline{31},\uline{37}\}$,$\{27\}$ \\
\cline{6-7}
&    &    &     &     & \color{blue!90} $\{\uline{34},\uline{37}\}$,$\{35\}$,$\{27\}$,$\{32\}$   & \color{blue!90} $\{\uline{34},\uline{37}\}$,$\{35\}$,$\{27\}$,$\{32\}$   \\
\cline{6-7}
&    &    &     &     & \multirow{7}{*}{$\{\uline{34},\uline{48}\}$,$\{35\}$,$\{32\}$,$\{27\}$}   &  $\{\uline{34},\uline{48}\}$,$\{35\}$,$\{32\}$,$\{27\}$   \\
\cline{7-7}
&    &    &     &     &    & \color{blue!90} $\{\uline{28},\uline{31},\uline{37}\}$,$\{29\}$,$\{27\}$  \\
\cline{7-7}
&    &    &     &     &    & \color{blue!90} $\{\uline{28},\uline{31},\uline{52}\}$,$\{29\}$,$\{27\}$ \\
\cline{7-7}
&    &    &     &     &    & \color{blue!90} $\{\uline{28},\uline{34},\uline{37}\}$,$\{35\}$,$\{27\}$,$\{29\}$,$\{32\}$ \\
\cline{7-7}
&    &    &     &     &    & \color{blue!90} $\{\uline{28},\uline{34},\uline{48}\}$,$\{35\}$,$\{27\}$,$\{29\}$,$\{32\}$ \\
\cline{7-7}
&    &    &     &     &    & \color{blue!90} $\{\uline{31},\uline{34},\uline{37}\}$,$\{27\}$,$\{35\}$ \\
\cline{7-7}
&    &    &     &     &    & \color{blue!90} $\{\uline{31},\uline{34},\uline{61}\}$,$\{27\}$,$\{35\}$ \\
\hline
\end{tabular}
\end{tiny}
\end{center}
\vspace{-0.5cm}
\caption{{\footnotesize MCSs et déviations identifiés  par {\tt LocFaults} pour des programmes sans boucles, sans l'usage du marquage des n\oe{}uds}}
\vspace{-0.7cm}

\label{MCSeskCF}
\end{table*}

\begin{table}[!h]
\begin{center}
\begin{scriptsize}
\begin{tabular}{|c|c|c|c|c|c|}
\hline
\multirow{3}{*}{Programme} & \multicolumn{5}{|c|}{LocFaults}  \\
\cline{2-6} &  \multirow{2}{*}{P}  &  \multicolumn{4}{|c|}{L}  \\
\cline{3-6}  &  & $= 0$ & $\leq 1$ & $\leq 2$ & $\leq 3$     \\
\hline
TritypeKO2  & $0,471$ & $0,023$  & $0,241$ & $2,529$ & $5,879$ \\
\hline
TritypeKO4 & $0,476$ & $0,022$ & $0,114$ & $0,348$ & $5,55$  \\
\hline
TriPerimetreKO3 & $0,487$  & $0,052$ &  $0,237$  & $2,468$ & $6,103$  \\
\hline
\end{tabular}
\end{scriptsize}
\end{center}
\vspace{-0.5cm}
\caption{{\footnotesize Temps de calcul, pour les résultats sans l'usage du marquage des n\oe{}uds}}
\label{time1}
\end{table}

\begin{table*}
\begin{center}
\begin{tiny}
\begin{tabular}{|c|c|c|c|c|c|c|}
\hline
\multirow{2}{*}{Programme} & \multirow{2}{*}{Contre-exemple} & \multirow{2}{*}{Erreurs} & \multicolumn{4}{|c|}{LocFaults} \\ 
\cline{4-7}
  &   &   & $= 0$ & $\leq 1$ & $\leq 2$ & $\leq 3$  \\
\hline
\multirow{7}{*}{TritypeKO2}  &  \multirow{7}{*}{$\{i=2,j=2,k=4\}$}  &  \multirow{7}{*}{$53$}  &  \multirow{7}{*}{$\{54\}$}  &   $\{54\}$  & $\{54\}$    &  $\{54\}$   \\ 
\cline{6-7}
\cline{5-5}
&    &    &     & $\{\uline{21}\}$    & $\{\uline{21}\}$   &  $\{\uline{21}\}$   \\
\cline{6-7}
\cline{5-5}
&    &    &     & $\{\uline{26}\}$  & $\{\uline{26}\}$   &  $\{\uline{26}\}$  \\
\cline{5-7}
&    &    &     &  $\{\uline{35}\}$,$\{27\}$,$\{25\}$   & $\{\uline{35}\}$,$\{27\}$,$\{25\}$   & $\{\uline{35}\}$,$\{27\}$,$\{25\}$  \\
\cline{6-7}
\cline{5-5}
&    &    &     & \multirow{3}{*}{$\{\uline{\textcolor{red}{53}}\}$,$\{25\}$,$\{27\}$}  &   $\{\uline{\textcolor{red}{53}}\}$,$\{25\}$,$\{27\}$  & $\{\uline{\textcolor{red}{53}}\}$,$\{25\}$,$\{27\}$       \\
\cline{6-7}
&    &    &     &     & $\{\uline{29},\uline{57}\}$,$\{30\}$,$\{27\}$,$\{25\}$   &  $\{\uline{29},\uline{57}\}$,$\{30\}$,$\{27\}$,$\{25\}$    \\
\cline{6-7}
&    &    &     &     &  $\{\uline{32},\uline{44}\}$,$\{33\}$,$\{25\}$, $\{27\}$  &  $\{\uline{32},\uline{44}\}$,$\{33\}$,$\{25\}$, $\{27\}$    \\   
\hline
\multirow{7}{*}{TritypeKO4}  &  \multirow{7}{*}{$\{i=2,j=3,k=3\}$}  &  \multirow{7}{*}{$45$}  &  \multirow{7}{*}{$\{46\}$}  &  $\{46\}$   & $\{46\}$   &  $\{46\}$    \\
\cline{5-7}
&    &    &     &   \multirow{6}{*}{$\{\uline{\textcolor{red}{45}}\}$,$\{33\}$,$\{25\}$}  & $\{\uline{\textcolor{red}{45}}\}$,$\{33\}$,$\{25\}$   & $\{\uline{\textcolor{red}{45}}\}$,$\{33\}$,$\{25\}$  \\
\cline{6-7}
&    &    &     &     & $\{\uline{26},\uline{32}\}$   &  $\{\uline{26},\uline{32}\}$  \\
\cline{6-7}
&    &    &     &     &  \multirow{4}{*}{$\{\uline{29},\uline{32}\}$}  & $\{\uline{29},\uline{32}\}$   \\
\cline{7-7}
&    &    &     &     &    & $\{\uline{32},\uline{35},\uline{49}\}$,$\{25\}$    \\
\cline{7-7}
&    &    &     &     &    & $\{\uline{32},\uline{35},\uline{53}\}$,$\{25\}$      \\
\cline{7-7}
&    &    &     &     &    & $\{\uline{32},\uline{35},\uline{57}\}$,$\{25\}$     \\
\hline
\multirow{7}{*}{TriPerimetreKO3}  &  \multirow{7}{*}{$\{i=2,j=1,k=2\}$}  &  \multirow{7}{*}{$57$}   &  \multirow{7}{*}{$\{58\}$}  &  $\{58\}$  & $\{58\}$  &  $\{58\}$   \\
\cline{5-7}
&    &    &     & $\{\uline{22}\}$  &  $\{\uline{22}\}$   &  $\{\uline{22}\}$    \\
\cline{5-7}
&    &    &     &  $\{\uline{31}\}$  & $\{\uline{31}\}$  &  $\{\uline{31}\}$   \\
\cline{5-7}
&    &    &     &  $\{\uline{37}\}$,$\{32\}$,$\{27\}$   &  $\{\uline{37}\}$,$\{32\}$,$\{27\}$   &   $\{\uline{37}\}$,$\{32\}$,$\{27\}$  \\
\cline{5-7}
&    &    &     &  \multirow{3}{*}{$\{\textcolor{red}{\uline{57}}\}$,$\{32\}$,$\{27\}$}  &  $\{\textcolor{red}{\uline{57}}\}$,$\{32\}$,$\{27\}$  & $\{\textcolor{red}{\uline{57}}\}$,$\{32\}$,$\{27\}$   \\
\cline{6-7}
&    &    &     &     & $\{\uline{28},\uline{61}\}$,$\{32\}$,$\{27\}$,$\{29\}$   &  $\{\uline{28},\uline{61}\}$,$\{32\}$,$\{27\}$,$\{29\}$    \\  
\cline{6-7}
&    &    &     &     & $\{\uline{34},\uline{48}\}$,$\{35\}$,$\{32\}$,$\{27\}$   &  $\{\uline{34},\uline{48}\}$,$\{35\}$,$\{32\}$,$\{27\}$   \\
\hline
\end{tabular}
\end{tiny}
\end{center}
\vspace{-0.5cm}
\caption{{\footnotesize MCSs et DCMs identifiés  par {\tt LocFaults} pour des programmes sans boucles, avec l'usage du marquage des n\oe{}uds}}
\vspace{-0.7cm}
\label{MCSeskCFmarke}
\end{table*}

\begin{table}[!h]
\begin{center}
\begin{scriptsize}
\begin{tabular}{|c|c|c|c|c|c|}
\hline
\multirow{3}{*}{Programme} & \multicolumn{5}{|c|}{LocFaults}  \\
\cline{2-6} &  \multirow{2}{*}{P}  &  \multicolumn{4}{|c|}{L}  \\
\cline{3-6}  &  & $= 0$ & $\leq 1$ & $\leq 2$ & $\leq 3$    \\
\hline
TritypeKO2  & $0,496$ & $0,022$  & $0,264$ & $\textbf{1,208}$ & $\textbf{1,119}$  \\
\hline
TritypeKO4 & $0,481$ & $0,021$ & $0,106$ & $\textbf{0,145}$ & $\textbf{1,646}$  \\
\hline
TriPerimetreKO3 & $0,485$  & $0,04$ &  $0,255$  & $\textbf{1,339}$ & $\textbf{1,219}$  \\
\hline
\end{tabular}
\end{scriptsize}
\end{center}
\vspace{-0.5cm}
\caption{{\footnotesize Temps de calcul, pour les résultats avec l'usage du marquage des n\oe{}uds}}
\label{time1marke}
\end{table}

\subsection{Les benchmarks avec boucles}
Ces benchmarks servent à mesurer l'extensibilité de {\tt LocFaults} par rapport à  {\tt BugAssist} pour des programmes avec boucles, en fonction de l'augmentation du nombre de dépliage $b$. Nous avons pris trois programmes avec boucles : BubbleSort, Sum et SquareRoot. Nous avons provoqué le bug \textit{Off-by-one} dans chacun. Le benchmark, pour chaque programme, est créé en faisant augmenter le nombre de dépliage $b$. $b$ est égal au nombre d'itérations effectuées par la boucle dans le pire des cas. Nous faisons aussi varier le nombre de conditions déviées pour {\tt LocFaults} de $0$ à $3$.

Nous avons utilisé le solveur MIP de CPLEX pour BubbleSort. Pour Sum et SquareRoot,  nous avons fait collaborer les deux solveurs de CPLEX (CP et MIP) lors du processus de la localisation. En effet, lors de la collecte des contraintes, nous utilisons une variable pour garder l'information sur le type du CSP construit. Quand {\tt LocFaults} détecte un chemin erroné\footnote{Un chemin erroné est celui sur lequel nous identifions les MCSs.} et avant de procéder au calcul des MCSs, il prend le bon solveur selon le type du CSP qui correspond à ce chemin : s'il est non linéaire, il utilise le solveur CP OPTIMIZER ; sinon, il utilise le solveur MIP.

Pour chaque benchmark, nous avons présenté un extrait de la table contenant les temps de calcul (les colonnes $P$ et $L$ affichent respectivement les temps de prétraitement et de calcul des MCSs), ainsi que le graphe qui correspond au temps de calcul des MCSs.
 
\subsubsection{Le benchmark BubbleSort}
BubbleSort est une implémentation de l'algorithme de tri à bulles. Ce programme contient deux boucles imbriquées ; sa complexité en moyenne est d'ordre $n^2$, où $n$ est la taille du tableau : le tri à bulles est considéré parmi les mauvais algorithmes de tri. L'instruction erronée dans ce programme entraîne le programme à trier le tableau en entrée en considérant seulement ses $n-1$ premiers éléments. Le mauvais fonctionnement du BubbleSort est dû au nombre d'itérations insuffisant effectué par la boucle. Cela est dû à l'initialisation fautive de la variable i :  i = tab.length - 1 ; l'instruction devait être i = tab.length. 
\begin{table}[!h]
{\fontsize{1pt}{1pt}\selectfont
\tabcolsep=2pt
\begin{center}
\begin{scriptsize}
\begin{tabular}{|c|c|c|c|c|c|c|c|c|}
\hline
\multirow{3}{*}{Programs} & \multirow{3}{*}{b} & \multicolumn{5}{|c|}{LocFaults} & \multicolumn{2}{c|}{BugAssist} \\
\cline{3-9} & & \multirow{2}{*}{P}  &  \multicolumn{4}{c|}{L} & \multirow{2}{*}{P} & \multirow{2}{*}{L} \\
\cline{4-7}  & & & $= 0$ & $\leq 1$ & $\leq 2$ & $\leq 3$ &   &    \\
\hline
V0 & $4$ & $ 0.751$  & $ 0.681$  & $ 0.56$  & $ 0.52$  & $ 0.948$  & $0.34$  & $55.27$ \\
\hline
V1 & $5$ & $ 0.813$  & $ 0.889$  & $ 0.713$  & $ 0.776$  & $ 1.331$  & $0.22$  & $125.40$ \\
\hline
V2 & $6$ & $ 1.068$  & $ 1.575$  & $ 1.483$  & $ 1.805$  & $ 4.118$  & $0.41$  & $277.14$ \\
\hline
V3 & $7$ & $ 1.153$  & $ 0.904$  & $ 0.85$  & $ 1.597$  & $ 12.67$  & $0.53$  & $612.79$ \\
\hline
V4 & $8$ & $ 0.842$  & $ 6.509$  & $ 6.576$  & $ 8.799$  & $ 116.347$  & $1.17$  & $1074.67$ \\
\hline
V5 & $9$ & $ 1.457$  & $ 18.797$  & $ 18.891$  & $ 21.079$  & $ 492.178$  & $1.24$  & $1665.62$ \\
\hline
V6 & $10$ & $ 0.941$  & $ 28.745$  & $ 29.14$  & $ 35.283$  & $ 2078.445$  & $1.53$  & $2754.68$ \\
\hline
V7 & $11$ & $ 0.918$  & $ 59.894$  & $ 65.289$  & $ 74.93$  & $ 4916.434$  & $3.94$  & $7662.90$ \\
\hline

\end{tabular}
\end{scriptsize}
\end{center}
\caption{Le temps de calcul pour le benchmark BubbleSort}
\label{TimesLB}
}
\end{table}

\begin{figure}[!h]
\begin{tikzpicture} 

\begin{axis}
[ 
  xlabel=Dépliages (b), ylabel=Temps (en secondes),ymin=0,xmin=0,
  height=9cm, width=7cm, grid=major,legend style={at={(1.20,0.99)}},
] 
\addplot coordinates { 
 
(4, 0.561)
(5, 0.597)
(6, 1.461)
(7, 0.813)
(8, 4.787)
(9, 14.234)
(10, 27.389)
(11, 56.008)
(12, 126.439)
(13, 235.282)
(14, 363.627)
(15, 437.994)
(16, 591.28)
(17, 737.541)
(18, 954.475)
(19, 1230.099)
(20, 1647.91)
};
\addlegendentry{LocFaults ($= 0$)} 

\addplot coordinates { 
  
(4, 0.553)
(5, 0.627)
(6, 1.496)
(7, 0.852)
(8, 4.911)
(9, 14.228)
(10, 27.608)
(11, 62.198)
(12, 126.233)
(13, 244.805)
(14, 360.651)
(15, 438.549)
(16, 621.072)
(17, 739.541)
(18, 1023.731)
(19, 1305.219)
(20, 1750.644)
}; 
\addlegendentry{LocFaults ($\leq 1$)} 

\addplot coordinates { 
  
(4, 0.508)
(5, 0.762)
(6, 1.75)
(7, 1.468)
(8, 6.01)
(9, 16.753)
(10, 33.573)
(11, 69.591)
(12, 157.238)
(13, 282.796)
(14, 500.626)
(15, 715.594)
(16, 971.357)
(17, 1726.373)
(18, 2197.53)
(19, 3477.862)
}; 
\addlegendentry{LocFaults ($\leq 2$)}

\addplot coordinates { 
  
(4, 0.948)
(5, 1.331)
(6, 4.118)
(7, 12.67)
(8, 116.347)
(9, 492.178)
(10, 2078.445)
(11, 4916.434)
}; 
\addlegendentry{LocFaults ($\leq 3$)}

\addplot coordinates { 
  
(4,55.27)
(5,125.40)
(6,277.14)
(7,612.79)
(8,1074.67)
(9,1665.62)
(10,2754.68)
(11,7662.90)
};
\addlegendentry{BugAssist}
\end{axis} 
\end{tikzpicture}
\caption{Comparaison de l'évolution des temps des différentes versions de {\tt LocFaults} et de {\tt BugAssist} pour le benchmark BubbleSort, en faisant augmenter le nombre d'itérations en dépliant la boucle.}
\label{LFvsBA}
\end{figure}
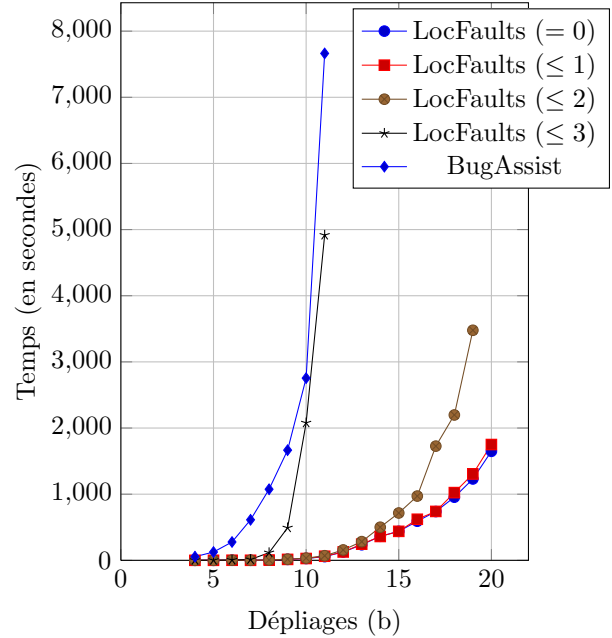
Les temps de {\tt LocFaults} et {\tt BugAssist} pour le benchmark BubbleSort sont présentés dans la table~\ref{TimesLB}. Le graphe qui illustre l'augmentation des temps des différentes versions de {\tt LocFaults} et de {\tt BugAssist} en fonction du nombre de dépliages est donné dans la figure~\ref{LFvsBA}.

La durée d'exécution de {\tt LocFaults} et de {\tt BugAssist} croît exponentiellement avec le nombre de dépliages ; les temps de {\tt BugAssist} sont toujours les plus grands. On peut considérer que {\tt BugAssist} est inefficace pour ce benchmark. Les différentes versions de {\tt LocFaults} (avec au plus $3$, $2$, $1$ et $0$ conditions déviées) restent utilisables jusqu'à un certain dépliage. Le nombre de dépliage au-delà de lequel la croissance des temps de {\tt BugAssist} devient rédhibitoire est inférieur à celui de {\tt LocFaults}, celui de {\tt LocFaults} avec au plus $3$ conditions déviées est inférieur à celui de {\tt LocFaults} avec au plus $2$ conditions déviées qui est inférieur lui aussi à celui de {\tt LocFaults} avec au plus $1$ conditions déviées. Les temps de {\tt LocFaults} avec au plus $1$ et $0$ condition déviée sont presque les mêmes.

\subsubsection{Les benchmarks SquareRoot et Sum}
Le programme SquareRoot (voir fig.~\ref{SquareRoot}) permet de trouver la partie entière de la racine carrée du nombre entier 50. Une erreur est injectée à la ligne 13, qui entraîne de retourner la valeur 8 ; or le programme doit retourner 7. Ce programme a été utilisé dans le papier décrivant l'approche {\tt BugAssist}, il contient un calcul numérique linéaire dans sa boucle et non linéaire dans sa postcondition.

\begin{figure}[!h]
    \center
\begin{lstlisting}{}
class SquareRoot{
  /*@ ensures((res*res<=val) && (res+1)*(res+1)>val);*/
  int SquareRoot()
    {
        int val = 50;
        int i = 1;
        int v = 0;
        int res = 0;
        while (v < val){
             v = v + 2*i + 1;
             i= i + 1; 
        }
        res = i; /*error: the instruction should be res = i - 1*/         
        return res;
   }
}
\end{lstlisting}
\vspace{-0.9cm}
\caption{Le programme SquareRoot}
\label{SquareRoot}
\end{figure}

Avec un dépliage égal à 50, {\tt BugAssist} calcule pour ce programme les instructions suspectes suivantes : $\{9, 10, 11, 13\}$. Le temps de la localisation est $36,16s$ et le temps de prétraitement est $0,12s$.

{\tt LocFaults} présente une instruction suspecte en indiquant à la fois son emplacement dans le programme (la ligne d'instruction), ainsi que la ligne de la condition et l'itération de chaque boucle menant à cette instruction. Par exemple, $9:2.11$ correspond à l'instruction qui se trouve à la ligne $11$ dans le programme, cette dernière est dans une boucle dont la ligne de la condition d'arrêt est $9$ et le numéro d'itération est $2$. Les ensembles suspectés par {\tt LocFaults} sont fournis dans le tableau suivant.
\begin{footnotesize}
\begin{tabular}{|c|c|}
  \hline
  \textit{DCMs} & \textit{MCSs}  \\
  \hline
  $\emptyset$ & $\{5\}$,$\{6\}$,$\{9:1.11\}$, $\{9:2.11\}$,$\{9:3.11\}$,  \\
   & $\{9:4.11\}$,$\{9:5.11\}$,$\{9:6.11\}$,$\{9:7.11\}$,$\{\textcolor{red}{13}\}$  \\
  \hline            
  \multirow{2}{*}{$\{9:7\}$} & $\{5\}$,$\{6\}$,$\{7\}$,$\{9:1.10\}$,$\{9:2.10\}$,$\{9:3.10\}$,  \\ 
  & $\{9:4.10\}$,$\{9:5.10\}$, $\{9:6.10\}$,$\{9:1.11\}$, \\
  & $\{9:2.11\}$,$\{9:3.11\}$,$\{9:4.11\}$,$\{9:5.11\}$, $\{9:6.11\}$ \\ 
  \hline
\end{tabular}
\end{footnotesize}

Le temps de prétraitement est $0,769s$. Le temps écoulé lors de l'exploration du CFG et le calcul des MCS est $1,299s$. Nous avons étudié le temps de {\tt LocFaults} et {\tt BugAssist} des valeurs de $val$ allant de $10$ à $100$ (le nombre de dépliage $b$ employé est égal à $val$), pour étudier le comportement combinatoire de chaque outil pour ce programme.

\begin{table}[h]
{\fontsize{1pt}{1pt}\selectfont
\tabcolsep=2pt
\begin{center}
\begin{scriptsize}
\begin{tabular}{|c|c|c|c|c|c|c|c|c|}
\hline
\multirow{3}{*}{Programs} & \multirow{3}{*}{b} & \multicolumn{5}{|c|}{LocFaults} & \multicolumn{2}{c|}{BugAssist} \\
\cline{3-9} & & \multirow{2}{*}{P}  &  \multicolumn{4}{c|}{L} & \multirow{2}{*}{P} & \multirow{2}{*}{L} \\
\cline{4-7}  & & & $= 0$ & $\leq 1$ & $\leq 2$ & $\leq 3$ &   &    \\
\hline
V0 & $10$ & $ 1.096$  & $ 1.737 $  & $ 2.098 $  & $ 2.113 $  & $ 2.066 $  & $0.05$  & $3.51$ \\
\hline
V10 & $20$ & $ 0.724$  & $ 0.974 $  & $ 1.131 $  & $ 1.117 $  & $ 1.099 $  & $0.05$  & $6.54$ \\
\hline
V20 & $30$ & $ 0.771$  & $ 1.048 $  & $ 1.16 $  & $ 1.171 $  & $ 1.223 $  & $0.08$  & $12.32$ \\
\hline
V30 & $40$ & $ 0.765$  & $ 1.048 $  & $ 1.248 $  & $ 1.266 $  & $ 1.28 $  & $0.09$  & $23.35$ \\
\hline
V40 & $50$ & $ 0.769$  & $ 1.089 $  & $ 1.271 $  & $ 1.291 $  & $ 1.299 $  & $0.12$  & $36.16$ \\
\hline
V50 & $60$ & $ 0.741$  & $ 1.041 $  & $ 1.251 $  & $ 1.265 $  & $ 1.281 $  & $0.14$  & $38.22$ \\
\hline
V70 & $80$ & $ 0.769$  & $ 1.114 $  & $ 1.407 $  & $ 1.424 $  & $ 1.386 $  & $0.19$  & $57.09$ \\
\hline
V80 & $90$ & $ 0.744$  & $ 1.085 $  & $ 1.454 $  & $ 1.393 $  & $ 1.505 $  & $0.22$  & $64.94$ \\
\hline
V90 & $100$ & $ 0.791$  & $ 1.168 $  & $ 1.605 $  & $ 1.616 $  & $ 1.613 $  & $0.24$  & $80.81$ \\
\hline

\end{tabular}
\end{scriptsize}
\end{center}
\caption{Le temps de calcul pour le benchmark SquareRoot}
\label{TableLFvsBASquareRoot}
}
\end{table}

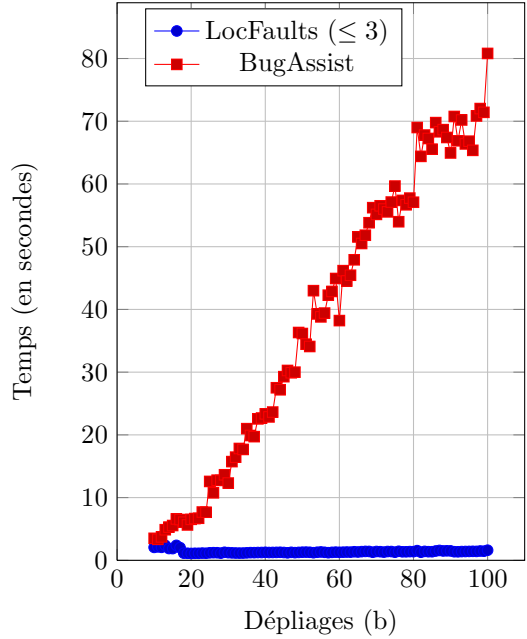
\begin{figure}[!h]
\begin{tikzpicture} 

\begin{axis}
[ 
  xlabel=Dépliages (b), ylabel=Temps (en secondes),ymin=0,xmin=0,
  height=9cm, width=7cm, grid=major,legend style={at={(0.70,0.99)}},
] 

\addplot coordinates { 
  
(10, 2.066 )
(11, 2.15 )
(12, 2.074 )
(13, 2.384 )
(14, 1.881 )
(15, 1.871 )
(16, 2.386 )
(17, 2.024 )
(18, 1.149 )
(19, 1.128 )
(20, 1.099 )
(21, 1.124 )
(22, 1.124 )
(23, 1.154 )
(24, 1.132 )
(25, 1.199 )
(26, 1.22 )
(27, 1.212 )
(28, 1.147 )
(29, 1.282 )
(30, 1.223 )
(31, 1.19 )
(32, 1.166 )
(33, 1.155 )
(34, 1.16 )
(35, 1.194 )
(36, 1.226 )
(37, 1.222 )
(38, 1.243 )
(39, 1.249 )
(40, 1.28 )
(41, 1.247 )
(42, 1.257 )
(43, 1.269 )
(44, 1.286 )
(45, 1.251 )
(46, 1.199 )
(47, 1.277 )
(48, 1.226 )
(49, 1.265 )
(50, 1.299 )
(51, 1.31 )
(52, 1.286 )
(53, 1.217 )
(54, 1.291 )
(55, 1.336 )
(56, 1.281 )
(57, 1.218 )
(58, 1.275 )
(59, 1.307 )
(60, 1.281 )
(61, 1.301 )
(62, 1.321 )
(63, 1.306 )
(64, 1.368 )
(65, 1.328 )
(66, 1.386 )
(67, 1.423 )
(68, 1.417 )
(69, 1.292 )
(70, 1.408 )
(71, 1.378 )
(72, 1.31 )
(73, 1.404 )
(74, 1.401 )
(75, 1.297 )
(76, 1.443 )
(77, 1.361 )
(78, 1.361 )
(79, 1.391 )
(80, 1.386 )
(81, 1.545 )
(82, 1.293 )
(83, 1.427 )
(84, 1.376 )
(85, 1.373 )
(86, 1.487 )
(87, 1.576 )
(88, 1.494 )
(89, 1.518 )
(90, 1.505 )
(91, 1.374 )
(92, 1.361 )
(93, 1.397 )
(94, 1.41 )
(95, 1.422 )
(96, 1.437 )
(97, 1.425 )
(98, 1.488 )
(99, 1.456 )
(100, 1.613 )
}; 
\addlegendentry{LocFaults ($\leq 3$)}

\addplot coordinates { 
  
(10,3.51)
(11,3.35)
(12,3.74)
(13,4.88)
(14,5.30)
(15,5.55)
(16,6.60)
(17,6.50)
(18,6.08)
(19,5.64)
(20,6.54)
(21,6.66)
(22,6.71)
(23,7.70)
(24,7.67)
(25,12.58)
(26,10.76)
(27,12.78)
(28,12.74)
(29,13.64)
(30,12.32)
(31,15.74)
(32,16.45)
(33,17.83)
(34,17.68)
(35,20.99)
(36,19.90)
(37,19.72)
(38,22.55)
(39,22.69)
(40,23.35)
(41,22.86)
(42,23.64)
(43,27.51)
(44,27.19)
(45,29.28)
(46,30.27)
(47,29.90)
(48,30.00)
(49,36.32)
(50,36.16)
(51,34.46)
(52,34.09)
(53,42.99)
(54,39.28)
(55,38.81)
(56,39.42)
(57,42.27)
(58,42.87)
(59,44.93)
(60,38.22)
(61,46.18)
(62,44.53)
(63,45.45)
(64,47.91)
(65,51.55)
(66,50.51)
(67,51.83)
(68,53.82)
(69,56.22)
(70,55.15)
(71,56.52)
(72,56.00)
(73,55.55)
(74,57.12)
(75,59.67)
(76,53.98)
(77,57.38)
(78,56.68)
(79,57.75)
(80,57.09)
(81,69.00)
(82,64.40)
(83,67.80)
(84,67.25)
(85,65.55)
(86,69.78)
(87,68.32)
(88,68.66)
(89,67.41)
(90,64.94)
(91,70.76)
(92,66.88)
(93,70.20)
(94,66.38)
(95,66.82)
(96,65.36)
(97,70.84)
(98,72.02)
(99,71.42)
(100,80.81)
};
\addlegendentry{BugAssist}

\end{axis} 
\end{tikzpicture}
\caption{Comparaison de l'évolution des temps de {\tt LocFaults} avec au plus $3$ conditions déviées et de {\tt BugAssist} pour le benchmark SquareRoot, en faisant augmenter le nombre d'itérations en dépliant la boucle.}
\label{GrapheLFvsBASquareRoot}
\end{figure}

Le programme Sum prend un entier positif $n$ de l'utilisateur, et il permet de calculer la valeur de $\sum_{i=1}^{n}i$. La postcondition spécifie cette somme. L'erreur dans Sum est dans la condition de sa boucle. Elle cause de calculer la somme $\sum_{i=1}^{n-1}i$ au lieu de $\sum_{i=1}^{n}i$. Ce programme contient des instructions numériques linéaires dans le c\oe{}ur de la boucle, et une postcondition non linéaire.  

\begin{table}[!h]
{\fontsize{1pt}{1pt}\selectfont
\tabcolsep=2pt
\begin{center}
\begin{scriptsize}
\begin{tabular}{|c|c|c|c|c|c|c|c|c|}
\hline
\multirow{3}{*}{Programs} & \multirow{3}{*}{b} & \multicolumn{5}{|c|}{LocFaults} & \multicolumn{2}{c|}{BugAssist} \\
\cline{3-9} & & \multirow{2}{*}{P}  &  \multicolumn{4}{c|}{L} & \multirow{2}{*}{P} & \multirow{2}{*}{L} \\
\cline{4-7}  & & & $= 0$ & $\leq 1$ & $\leq 2$ & $\leq 3$ &   &    \\
\hline
V0 & $6$ & $ 0.765$  & $ 0.427$  & $ 0.766$  & $ 0.547$  & $ 0.608$  & $0.04$  & $2.19$ \\
\hline
V10 & $16$ & $ 0.9$  & $ 0.785$  & $ 1.731$  & $ 1.845$  & $ 1.615$  & $0.08$  & $17.88$ \\
\hline
V20 & $26$ & $ 1.11$  & $ 1.449$  & $ 7.27$  & $ 7.264$  & $ 6.34$  & $0.12$  & $53.85$ \\
\hline
V30 & $36$ & $ 1.255$  & $ 0.389$  & $ 8.727$  & $ 4.89$  & $ 4.103$  & $0.13$  & $108.31$ \\
\hline
V40 & $46$ & $ 1.052$  & $ 0.129$  & $ 5.258$  & $ 5.746$  & $ 13.558$  & $0.23$  & $206.77$ \\
\hline
V50 & $56$ & $ 1.06$  & $ 0.163$  & $ 7.328$  & $ 6.891$  & $ 6.781$  & $0.22$  & $341.41$ \\
\hline
V60 & $66$ & $ 1.588$  & $ 0.235$  & $ 13.998$  & $ 13.343$  & $ 14.698$  & $0.36$  & $593.82$ \\
\hline
V70 & $76$ & $ 0.82$  & $ 0.141$  & $ 10.066$  & $ 9.453$  & $ 10.531$  & $0.24$  & $455.76$ \\
\hline
V80 & $86$ & $ 0.789$  & $ 0.141$  & $ 13.03$  & $ 12.643$  & $ 12.843$  & $0.24$  & $548.83$ \\
\hline
V90 & $96$ & $ 0.803$  & $ 0.157$  & $ 34.994$  & $ 28.939$  & $ 18.141$  & $0.31$  & $785.64$ \\
\hline
\end{tabular}
\end{scriptsize}
\end{center}
\caption{Le temps de calcul pour le benchmark Sum}
\label{TableLFvsBASum}
}
\end{table}

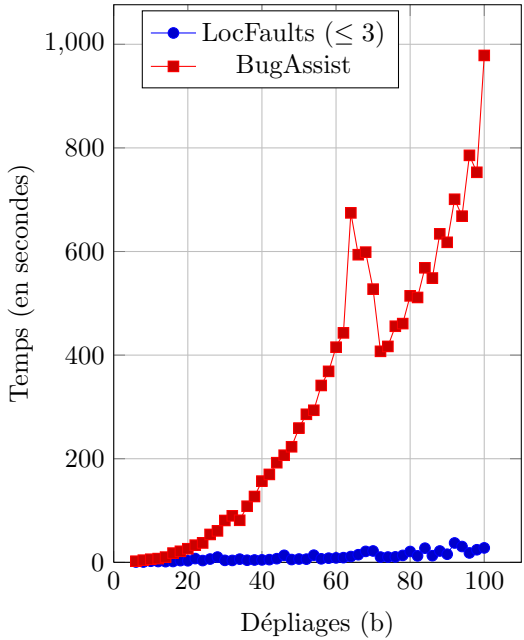
\begin{figure}[!h]
\begin{tikzpicture} 

\begin{axis}
[ 
  xlabel=Dépliages (b), ylabel=Temps (en secondes),ymin=0,xmin=0,
  height=9cm, width=7cm, grid=major,legend style={at={(0.70,0.99)}},
] 

\addplot coordinates { 
  
(6, 0.608)
(8, 0.583)
(10, 2.103)
(12, 1.567)
(14, 1.308)
(16, 1.615)
(18, 3.279)
(20, 3.0)
(22, 7.216)
(24, 3.436)
(26, 6.34)
(28, 10.17)
(30, 3.809)
(32, 3.681)
(34, 6.187)
(36, 4.103)
(38, 4.502)
(40, 4.706)
(42, 5.202)
(44, 7.047)
(46, 13.558)
(48, 5.396)
(50, 6.384)
(52, 6.163)
(54, 13.819)
(56, 6.781)
(58, 8.055)
(60, 8.553)
(62, 9.03)
(64, 11.205)
(66, 14.698)
(68, 21.123)
(70, 21.921)
(72, 10.175)
(74, 10.206)
(76, 10.531)
(78, 13.366)
(80, 20.915)
(82, 12.487)
(84, 27.336)
(86, 12.843)
(88, 21.9)
(90, 15.67)
(92, 37.395)
(94, 30.523)
(96, 18.141)
(98, 24.283)
(100, 27.856)
}; 
\addlegendentry{LocFaults ($\leq 3$)}

\addplot coordinates { 
  
(6,2.19)
(8,4.22)
(10,5.98)
(12,7.20)
(14,10.14)
(16,17.88)
(18,21.31)
(20,26.24)
(22,33.18)
(24,37.36)
(26,53.85)
(28,60.68)
(30,80.80)
(32,89.79)
(34,81.19)
(36,108.31)
(38,127.10)
(40,156.84)
(42,169.73)
(44,192.34)
(46,206.77)
(48,223.07)
(50,259.24)
(52,285.80)
(54,293.56)
(56,341.41)
(58,368.68)
(60,415.34)
(62,442.89)
(64,674.55)
(66,593.82)
(68,598.82)
(70,527.41)
(72,407.33)
(74,416.81)
(76,455.76)
(78,460.84)
(80,514.36)
(82,511.29)
(84,568.54)
(86,548.83)
(88,634.07)
(90,617.77)
(92,700.87)
(94,668.30)
(96,785.64)
(98,752.92)
(100,978.51)
};
\addlegendentry{BugAssist}

\end{axis} 
\end{tikzpicture}
\caption{Comparaison de l'évolution des temps de {\tt LocFaults} avec au plus $3$ conditions déviées et de {\tt BugAssist} pour le benchmark Sum, en faisant augmenter le nombre d'itérations en dépliant la boucle.}
\label{GrapheLFvsBASum}
\end{figure}

Les résultats en temps pour les benchmarks SquareRoot et Sum sont présentés dans les tables respectivement~\ref{TableLFvsBASquareRoot} et~\ref{TableLFvsBASum}. Nous avons dessiné aussi le graphe qui correspond au résultat de chaque benchmark, voir respectivement le graphe de la figure~\ref{GrapheLFvsBASquareRoot} et~\ref{GrapheLFvsBASum}. Le temps d'exécution de {\tt BugAssist} croît rapidement ; les temps de {\tt LocFaults} sont presque constants. Les temps de {\tt LocFaults} avec au plus $0$, $1$ et $2$ conditions déviées sont proches de ceux de {\tt LocFaults} avec au plus $3$ conditions déviées.

\section{Conclusion}
La méthode {\tt LocFaults} détecte les sous-ensembles suspects en analysant les chemins du CFG pour trouver les DCMs et les MCSs à partir de chaque DCM ; elle utilise des solveurs de contraintes. La méthode {\tt BugAssit} calcule la fusion des MCSs du programme en transformant le programme complet en une formule booléenne ; elle utilise des solveurs Max-SAT. Les deux méthodes travaillent en partant d'un contre-exemple. Dans ce papier, nous avons présenté une exploration de la scalabilité de {\tt LocFaults}, particulièrement sur le traitement des boucles avec le bug \textit{Off-by-one}. Les premiers résultats montrent que {\tt LocFaults} est plus efficace que {\tt BugAssist} sur des programmes avec boucles.
 Les temps de {\tt BugAssist} croissent rapidement en fonction du nombre de dépliages.
 
Dans le cadre de nos travaux futurs, nous envisageons de confirmer nos résultats sur des programmes avec boucles plus complexes. Nous développons une version interactive de notre outil qui fournit les sous-ensembles suspects l'un après l'autre : nous voulons tirer profit des connaissances de l'utilisateur pour sélectionner les conditions qui doivent être déviées. Nous réfléchissons également sur comment étendre notre méthode pour supporter les instructions numériques avec calcul sur les flottants.

\paragraph{Remerciements.} Nous remercions Bertrand Neveu pour sa lecture attentive et ses commentaires utiles sur ce papier. Merci également à Michel Rueher et Hélène Collavizza pour leurs remarques intéressantes.

\begin{small}

\end{small}
%\bibliography{jfpc15}

\end{document}